\documentclass[10pt,twocolumn,letterpaper]{article}

\usepackage{cvpr}
\usepackage{times}
\usepackage{epsfig}
\usepackage{graphicx}
\usepackage{amsmath}
\usepackage{amssymb}

\usepackage{mathrsfs}
\usepackage{url}
\usepackage{enumitem}
\usepackage{algorithm}
\usepackage{algpseudocode}
\usepackage{dsfont}
\usepackage{multirow}
\usepackage{booktabs}
\usepackage[pagebackref=false,breaklinks=true,letterpaper=true,colorlinks,urlcolor=blue,citecolor=blue,linkcolor=blue,bookmarks=false]{hyperref}

\usepackage{bbm}
\usepackage{amsmath,bm}

\cvprfinalcopy 

\def\cvprPaperID{2141} 

\begin{document}

\title{Transformer Meets Tracker: \\ Exploiting Temporal Context for Robust Visual Tracking}

\author{Ning~Wang$^{1}$~~~Wengang~Zhou$^{1,2}$~~~Jie~Wang$^{1,2}$~~~Houqiang~Li$^{1,2}$\\
	{\normalsize $^{1}$CAS Key Laboratory of GIPAS, EEIS Department, University of Science and Technology of China (USTC)} \\
    {\normalsize $^{2}$Institute of Artificial Intelligence, Hefei Comprehensive National Science Center} \\
	{\tt\small wn6149@mail.ustc.edu.cn, \{zhwg,jiewangx,lihq\}@ustc.edu.cn}
}

\maketitle

\newcommand\blfootnote[1]{%
\begingroup 
\renewcommand\thefootnote{}\footnote{#1}%
\addtocounter{footnote}{-1}%
\endgroup 
}

\begin{abstract}
	In video object tracking, there exist rich temporal contexts among successive frames, which have been largely overlooked in existing trackers.
	In this work, we bridge the individual video frames and explore the temporal contexts across them via a transformer architecture for robust object tracking.
	Different from classic usage of the transformer in natural language processing tasks, we separate its encoder and decoder into two parallel branches and carefully design them within the Siamese-like tracking pipelines.
	The transformer encoder promotes the target templates via attention-based feature reinforcement, which benefits the high-quality tracking model generation.
	The transformer decoder propagates the tracking cues from previous templates to the current frame, which facilitates the object searching process.
	Our transformer-assisted tracking framework is neat and trained in an end-to-end manner.
	With the proposed transformer, a simple Siamese matching approach is able to outperform the current top-performing trackers.
	By combining our transformer with the recent discriminative tracking pipeline, our method sets several new state-of-the-art records on prevalent tracking benchmarks. 
\end{abstract}

\blfootnote{*Corresponding Author: Wengang Zhou and Houqiang Li.} 
\blfootnote{\dag Source code, pretrained model, and raw tracking results are available at \url{https://github.com/594422814/TransformerTrack}. }

\vspace{-0.1in}
\section{Introduction}

Visual object tracking is a basic task in computer vision. Despite the recent progress, it remains a challenging task due to factors such as occlusion, deformation, and appearance changes.
With the temporal error accumulation, these challenges are further amplified in the online process.

It is well recognized that the rich temporal information in the video flow is of vital importance for visual tracking.
However, most tracking paradigms \cite{SiamRPN,siamrpn++,MAML} handle this task by per-frame object detection, where the temporal relationships among successive frames have been largely overlooked.
Take the popular Siamese tracker as an example, only the initial target is considered for template matching \cite{SiamFC,SINT,GOTURN,SiamRPN}.
The merely used temporal information is the motion prior (\emph{e.g.,} cosine window) by assuming the target moves smoothly, which is widely adopted in visual trackers.
In other tracking frameworks with update mechanisms \cite{KCF,MDNet,ECO,MemTrack,SiameseUpdate,DiMP}, previous prediction results are collected to incrementally update the tracking model.
Despite the historical frames considered in the above approaches, the video frames are still considered as independent counterparts without mutual reasoning.
In real-world videos, some frames inevitably contain noisy contents such as occluded or blurred objects.
These imperfect frames will hurt the model update when serving as the templates and will challenge the tracking process when performing as the search frames.  
Therefore, it is a non-trivial issue to convey rich information across temporal frames to mutually reinforce them.
We argue that the video frames should not be treated in isolation and the performance potential is largely restricted due to the overlook of frame-wise relationship.

\begin{figure}
	\centering
	\includegraphics[width=8.1cm]{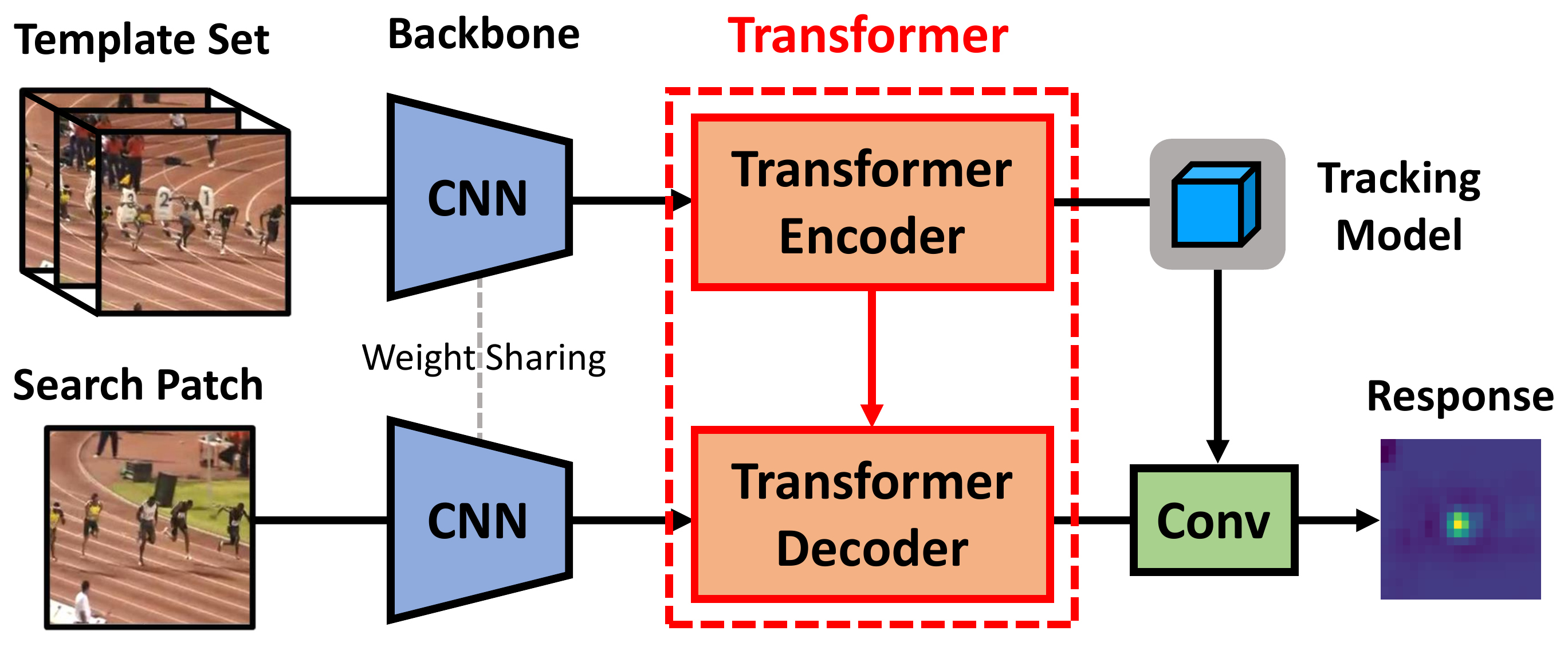}
	\caption{An overview of our transformer-assisted tracking framework. The transformer encoder and decoder are assigned to two parallel branches in a Siamese-like tracking pipeline. Thanks to the encoder-decoder structure, isolated frames are tightly bridged to convey rich temporal information in the video flow.}\label{fig:1}
	\vspace{-0.1in}
\end{figure}

To bridge the isolated video frames and convey the rich temporal cues across them, in this work, we introduce the transformer architecture \cite{Transformer} to the visual tracking community.
Different from the traditional usage of the transformer in language modeling and machine translation \cite{Transformer,BERT}, we leverage it to handle the \emph{context propagation in the temporal domain}.
By carefully modifying the classic transformer architecture, we show that its transformation characteristic naturally fits the tracking scenario.
Its core component, \emph{i.e.,} attention mechanism \cite{Transformer,nonlocal}, is ready to establish the pixel-wise correspondence across frames and freely convey various signals in the temporal domain.

Generally, most tracking methods \cite{SiamFC,CFNet,SiamRPN,CREST,ATOM,DiMP} can be formulated into a Siamese-like framework, where the top branch learns a tracking model using template features, and the bottom branch classifies the current search patch. 
As shown in Figure~\ref{fig:1}, we separate the transformer encoder and decoder into two branches within such a general Siamese-like structure. 
In the top branch, a set of template patches are fed to the transformer encoder to generate high-quality encoded features.
In the bottom branch, the search feature as well as the previous template contents are fed to the transformer decoder, where the search patch retrieves and aggregates informative target cues (\emph{e.g.,} spatial masks and target features) from history templates to reinforce itself.  
%

\begin{figure}
	\centering
	\includegraphics[width=8.2cm]{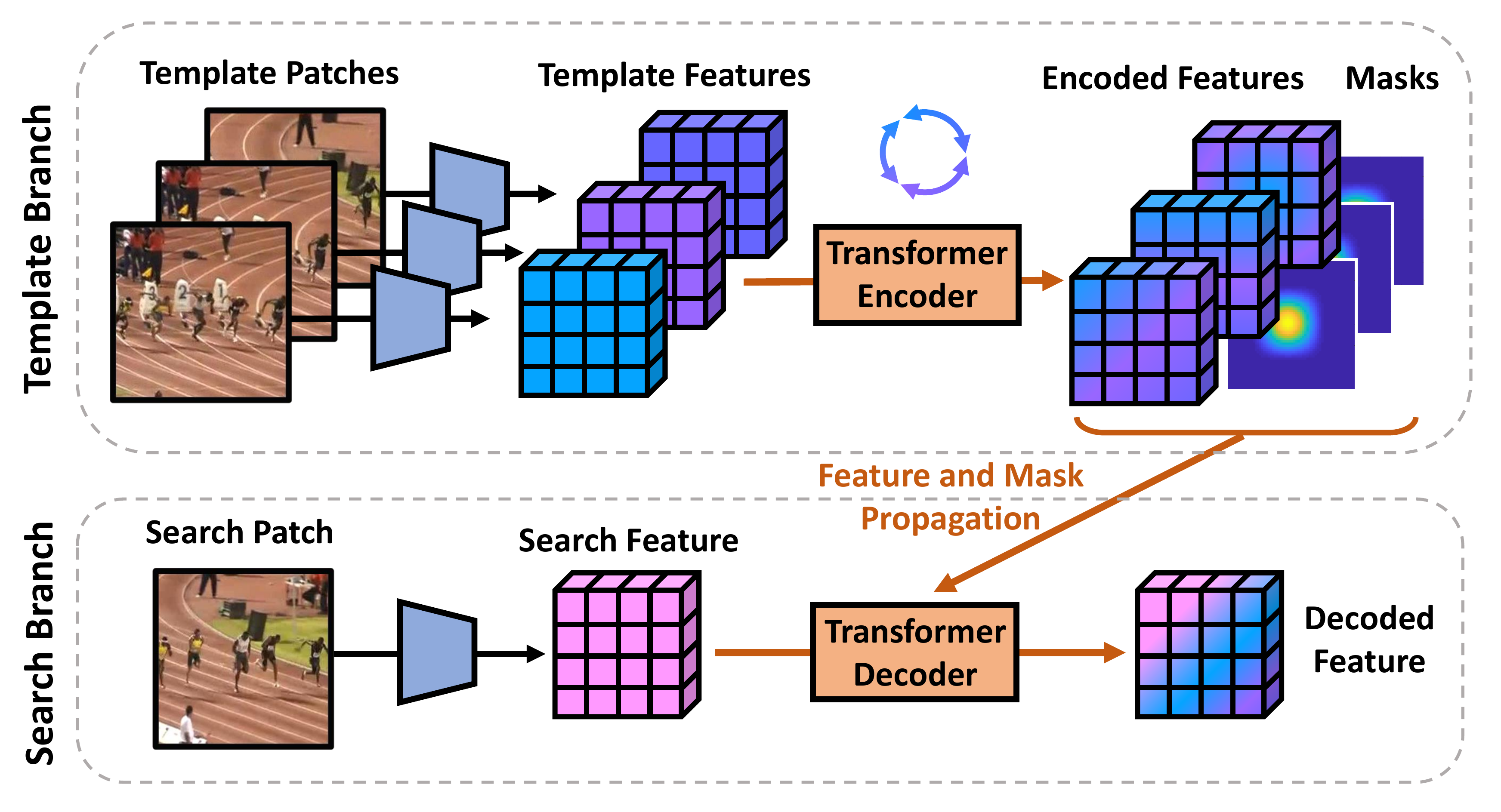}
	\caption{Top: the transformer encoder receives multiple template features to mutually aggregate representations. Bottom: the transformer decoder propagates the template features and their assigned masks to the search patch feature for representation enhancement.}\label{fig:2}
	\vspace{-0.1in}
\end{figure}

The proposed transformer facilitates visual tracking via:
\begin{itemize}[noitemsep,nolistsep]	
	\item {\bf Transformer Encoder.} It enables individual template features to mutually reinforce to acquire more compact target representations, as shown in Figure~\ref{fig:2}. 
	These encoded high-quality features further benefit the tracking model generation.
	
	\item {\bf Transformer Decoder.} It conveys valuable temporal information across frames. As shown in Figure~\ref{fig:2}, our decoder simultaneously transfers features and spatial masks.
	Propagating the features from previous frames to the current patch smooths the appearance changes and remedies the context noises while transforming the spatial attentions highlights the potential object location. 
	These manifold target representations and spatial cues make the object search much easier.
\end{itemize}
Finally, we track the target in the decoded search patch. 
To verify the generalization of our designed transformer, we integrate it into two popular tracking frameworks including a Siamese formulation \cite{SiamFC} and a discriminative correlation filter (DCF) based tracking paradigm \cite{DiMP}.
With our designed transformer, a simple Siamese matching pipeline is able to outperform the current top-performing trackers.
By combining with the recent discriminative approach \cite{DiMP}, our transformer-assisted tracker shows outstanding results on seven prevalent tracking benchmarks including LaSOT \cite{LaSOT}, TrackingNet \cite{2018trackingnet}, GOT-10k \cite{GOT10k}, UAV123 \cite{UAV123}, NfS \cite{NFSdataset}, OTB-2015 \cite{OTB-2015}, and VOT2018 \cite{VOT2018}  and sets several new state-of-the-art records.

In summary, we make three-fold contributions:
\begin{itemize}[noitemsep,nolistsep]	
	\item We present a neat and novel transformer-assisted tracking framework. To our best knowledge, this is the first attempt to involve the transformer in visual tracking.	
	
	\item We simultaneously consider the feature and attention transformations to better explore the potential of the transformer. We also modify the classic transformer to make it better suit the tracking task.	
	
	\item To verify the generalization, we integrate our designed transformer into two popular tracking pipelines. Our trackers exhibit encouraging results on 7 benchmarks. 
\end{itemize}

\section{Related Work}
{\noindent \bf Visual Tracking.} Given the initial target in the first frame, visual tracking aims to localize it in successive frames.
In recent years, the Siamese network has gained significant popularity, which deals with the tracking task by template matching \cite{SiamFC,SINT,GOTURN}. 
By introducing the region proposal network (RPN), Siamese trackers obtain superior efficiency and more accurate target scale estimation \cite{SiamRPN,DaSiamRPN}.
The recent improvements upon Siamese trackers include attention mechanism \cite{RASNet}, reinforcement learning \cite{EAST,POST}, target-aware model fine-tuning \cite{TADT}, unsupervised training \cite{UDT,LUDT}, sophisticated backbone networks \cite{siamrpn++,deeperwiderSiamFC}, cascaded frameworks \cite{CRPN,SPM}, and model update mechanisms \cite{GCT,Dsiam,MemTrack,SiameseUpdate}.

Discriminative correlation filter (DCF) tackles the visual tracking by solving the ridge regression in Fourier domain, which exhibits attractive efficiency \cite{KCF,HCF,CSR-DCF,BACF,Context-AwareCorrelationFilter,MCCT,C-COT,ECO}. 
%
%
The recent advances show that the ridge regression can be solved in the deep learning frameworks \cite{CREST,DSLT,ATOM,DiMP}, which avoids the boundary effect in classic DCF trackers.
These methods learn a discriminative CNN kernel to convolve with the search area for response generation.
In recent works, the residual terms \cite{CREST} and shrinkage loss \cite{DSLT} are incorporated into the deep DCF formulation.  
To accelerate the kernel learning process, ATOM \cite{ATOM} exploits the conjugate gradient algorithm. 
The recent DiMP tracker \cite{DiMP} enhances the discriminative capability of the learned CNN kernel in an end-to-end manner, which is further promoted by the probabilistic regression framework \cite{PrDiMP}.

Despite the impressive performance, most existing methods \cite{MDNet,SiamFC,SiamRPN,ATOM,DiMP,D3S,MAML} generally regard the tracking task as the per-frame object detection problem, failing to adequately exploit the temporal characteristic of the tracking task.
Some previous works explore the temporal information using graph neural network \cite{GCT}, spatial-temporal regularization \cite{STRCF}, optical flow \cite{FlowCF}, \emph{etc}. 
Differently, we leverage the transformer to model the frame-wise relationship and propagate the temporal cues, which is neat and ready to integrate with the modern deep trackers.
%

{\noindent \bf Transformer.} Transformer is first proposed in \cite{Transformer} as a new paradigm for machine translation.
The basic block in a transformer is the attention module, which aggregates information from the entire input sequence.
Due to the parallel computations and unique memory mechanism, transformer architecture is more competitive than RNNs in processing long sequences and has gained increasing popularity in many natural language processing (NLP) tasks \cite{BERT,language2019,end-to-endASR}.   
Similarly, non-local neural network \cite{nonlocal} also introduces a self-attention block to acquire global representations, which has been adopted in many vision tasks including visual object tracking \cite{DeformableSiam}.
Nevertheless, how to take advantage of the compact transformer encoder-decoder structure for visual tracking has been rarely studied.

Recently, transformer architecture has been introduced to computer vision such as image generation \cite{ImageTransformer}. 
Transformer based object detection approach is proposed in \cite{DETR}, which views the object detection task as a direct set prediction problem. 
However, the above techniques leverage the transformer in the image-level tasks.
In this paper, we show that the transformer structure serves as a good fit for video-related scenarios by transferring temporal information across frames.
%
%
To bridge the domain gap between visual tracking and NLP tasks, we carefully modify the classic transformer to better suit the tracking scenario.

\section{Revisting Tracking Frameworks}\label{revisiting tracking}
\begin{figure}
	\centering
	\includegraphics[width=8.2cm]{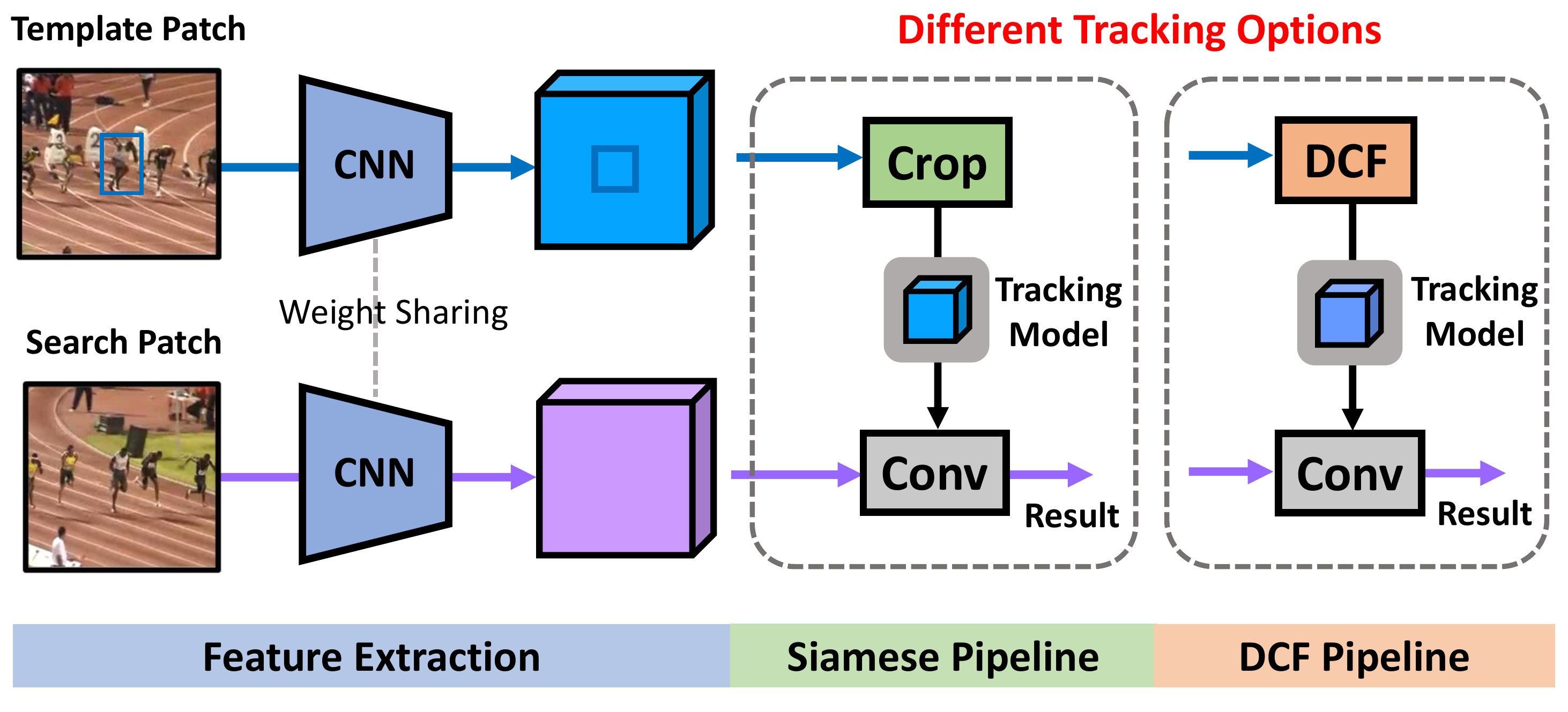}
	\caption{The simplified pipelines of Siamese \cite{SiamFC} and DCF \cite{ATOM,DiMP} based trackers. These tracking approaches can be formulated into a Siamese-like pipeline, where the top branch is responsible for the model generation and the bottom branch localizes the target.}\label{fig:tracking_methods}
	\vspace{-0.08in}
\end{figure}

Before elaborating our transformer for object tracking, we briefly review the recent popular tracking approaches for the sake of completeness.
As shown in Figure~\ref{fig:tracking_methods}, the mainstream tracking methods such as Siamese network \cite{SiamFC} or discriminative correlation filter (DCF) \cite{CFNet,ATOM,DiMP} can be formulated into the Siamese-like pipeline, where the top branch learns the tracking model using templates and the bottom branch focuses on the target localization.

Siamese matching architecture \cite{SiamFC} takes an exemplar patch $ \bm{z} $ and a search patch $ \bm{x} $ as inputs, where $ \bm{z} $ represents the target object while $ \bm{x} $ is a large searching area in subsequent video frames. 
Both of them are fed to the weight-sharing CNN network $ \Psi(\cdot) $. 
Their output feature maps are cross-correlated as follows to generate the response map:
\begin{equation}\label{EQ1}
\bm{r}(\bm{z}, \bm{x}) = \Psi(\bm{z}) \ast \Psi(\bm{x}) + b \cdot \mathbbm{1},
\end{equation}
where $ \ast $ is the cross-correlation and $ b \cdot \mathbbm{1} $ denotes a bias term.
Siamese trackers rely on the target model, \emph{i.e.,} convolutional kernel $\Psi(\bm{z}) $, for template matching.

As another popular framework, deep learning based DCF method optimizes the tracking model $ \bm{f} $ under a ridge regression formulation \cite{CREST,ATOM,DiMP} as follows:
\begin{equation}\label{EQ2}
\min_{ \bm{f}} \|\bm{f} \ast \Psi(\bm{z}^{\star}) - \bm{y}\|^2_2 + \lambda\|\bm{f}\|^2_2,
\end{equation}
where $ \bm{y} $ is the Gaussian-shaped ground-truth label of template patch $ \bm{z}^{\star} $, and $ \lambda $ controls the regularization term to avoid overfitting.
Note that $ \bm{z}^{\star} $ is much larger than the exemplar patch $ \bm{z} $  in Siamese trackers.
Therefore, DCF formulation simultaneously considers the target matching and background discrimination.
After obtaining the tracking model $ \bm{f} $, the response is generated via $ \bm{r} = \bm{f} \ast \Psi(\bm{x}) $.

The traditional DCF methods \cite{KCF,DSST} solve ridge regression using circularly generated samples via the closed-form solution in the Fourier domain.
In contrast, the recent deep learning based DCF methods solve Eq.~\ref{EQ2} using stochastic gradient descent \cite{CREST,DSLT} or conjugate gradient approach \cite{ATOM} to avoid the boundary effect.
The recent DiMP \cite{DiMP} optimizes the above ridge regression via a meta-learner in an end-to-end manner, showing state-of-the-art performance.

\section{Transformer for Visual Tracking}
As discussed in Section~\ref{revisiting tracking}, mainstream tracking methods can be formulated into a Siamese-like pipeline. 
We aim to improve such a general tracking framework by frame-wise relationship modeling and temporal context propagation, without modifying their original tracking manners such as template matching.
%
%

%
%

\subsection{Transformer Overview}\label{guidelines}

\begin{figure}
	\centering
	\includegraphics[width=8.2cm]{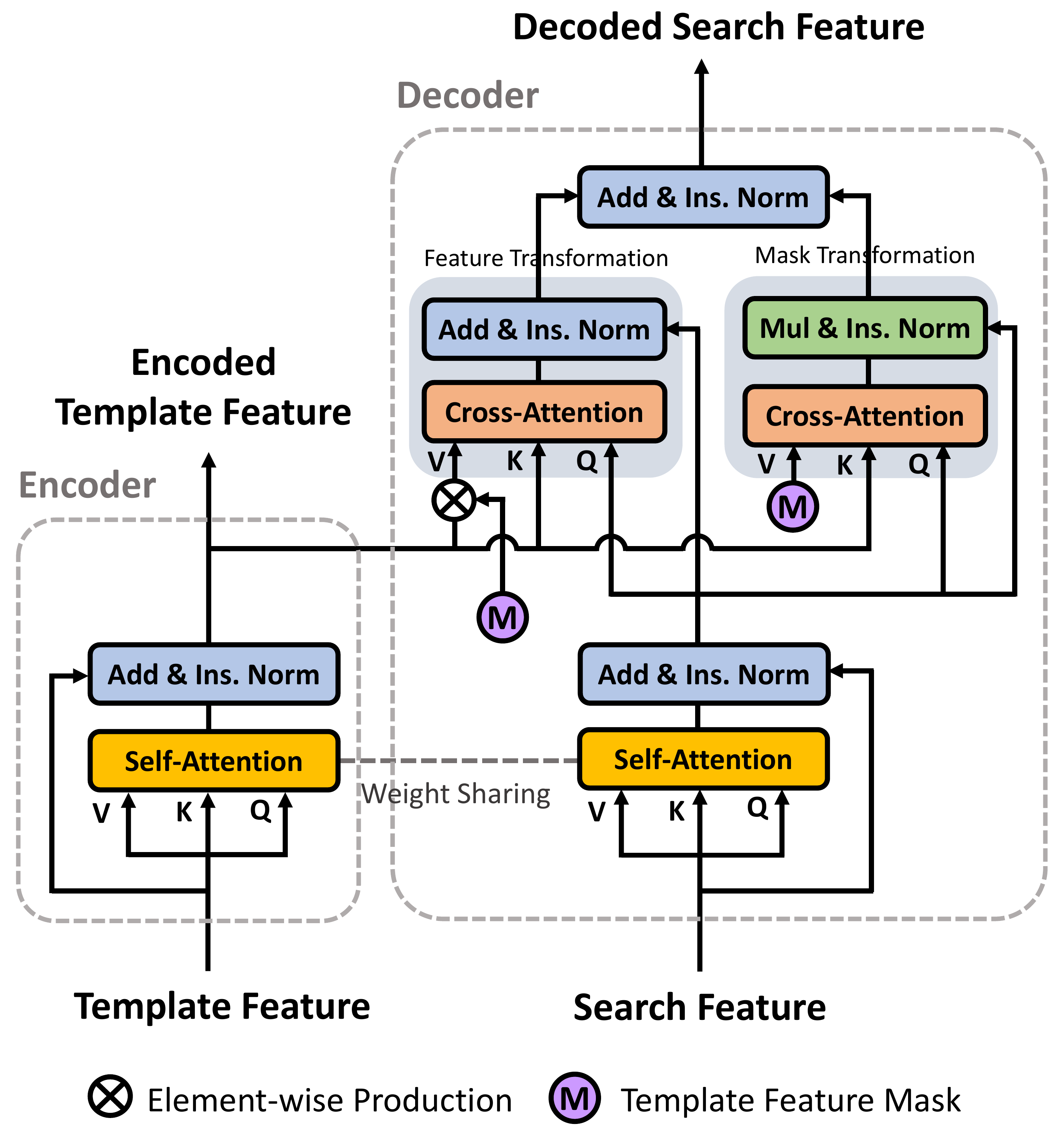}
	\caption{An overview of the proposed transformer architecture.}\label{fig:main}
	\vspace{-0.1in}
\end{figure}

An overview of our transformer is shown in Figure~\ref{fig:main}. Similar to the classic transformer architecture \cite{Transformer}, the encoder leverages self-attention block to mutually reinforce multiple template features. 
In the decoding process, cross-attention block bridges template and search branches to propagate temporal contexts (\emph{e.g.,} feature and attention).

%
To suit the visual tracking task, we modify the classic transformer in the following aspects:
(1) \emph{Encoder-decoder Separation.} Instead of cascading the encoder and decoder in NLP tasks \cite{Transformer,BERT}, as shown in Figure~\ref{fig:1}, we separate the encoder and decoder into two branches to fit the Siamese-like tracking methods.
(2) \emph{Block Weight-sharing.} The self-attention blocks in the encoder and decoder (yellow boxes in Figure~\ref{fig:main}) share weights, which transform the template and search embeddings in the same feature space to facilitate the further cross-attention computation.
(3) \emph{Instance Normalization.} In NLP tasks \cite{Transformer}, the word embeddings are individually normalized using the layer normalization.
Since our transformer receives image feature embeddings, we jointly normalize these embeddings in the instance (image patch) level to retain the valuable image amplitude information.
(4) \emph{Slimming Design.} Efficiency is crucial for visual tracking scenarios. To achieve a good balance of speed and performance, we slim the classic transformer by omitting the fully-connected feed-forward layers and maintaining the lightweight single-head attention.

\subsection{Transformer Encoder}\label{encoder}

The basic block in a classic transformer is the attention mechanism, which receives the query $ {\bf Q}\in \mathbb{R}^{ N_q \times C} $, key $ {\bf K} \in \mathbb{R}^{ N_k \times C} $, and value $ {\bf V} \in \mathbb{R}^{ N_k \times C} $ as the inputs.
In our approach, following \cite{Transformer}, we also adopt the dot-product to compute the similarity matrix $ {\bf A}_{\text{K}\to\text{Q}} \in \mathbb{R}^{ N_q \times N_k} $ between the query and key as follows: 
\begin{equation}\label{EQ3}
{\bf A}_{\text{K}\to\text{Q}} = \text{Atten}({\bf Q}, {\bf K}) = \text{Softmax}_{\text{col}}(\bar{\bf Q}\bar{\bf K}^{\mathrm{T}}/\tau),
\end{equation}
where $ \bar{\bf Q} $ and $ \bar{\bf K} $ are $ \ell_2 $-normalized features of $ \bf Q $ and $\bf K $ across the channel dimension, and $ \tau $ is a temperature parameter controlling the Softmax distribution, which is inspired by the model distillation \cite{hinton2015distilling} and contrastive learning \cite{SimCLR} techniques.
With the propagation matrix $ {\bf A}_{\text{K}\to\text{Q}} $ from key to query, we can transform the value via $ {\bf A}_{\text{K}\to\text{Q}}{\bf V} \in \mathbb{R}^{N_q \times C} $.

In our framework, the transformer encoder receives a set of template features $ {\bf T}_{i} \in \mathbb{R}^{C \times H \times W} $ with a spatial size of $ H \times W $ and dimensionality $ C $, which are further concatenated to form the template feature ensemble $ {\bf T} = \text{Concat}({\bf T}_{1}, \cdots, {\bf T}_{n}) \in \mathbb{R}^{n \times C \times H \times W} $.
To facilitate the attention computation, we reshape ${\bf T}$ to $ {\bf T}^{'} \in \mathbb{R}^{N_T \times C} $, where $ N_T=n\times H \times W $.
As shown in Figure~\ref{fig:main}, the main operation in the transformer encoder is self-attention, which aims to mutually reinforce the features from multiple templates.
To this end, we first compute the self-attention map $ {\bf A}_{\text{T}\to\text{T}} = \text{Atten}\left(\varphi({\bf T}^{'}), \varphi({\bf T}^{'}) \right)  \in \mathbb{R}^{N_T\times N_T}$,  where $ \varphi(\cdot) $ is a $ 1\times 1 $ linear transformation that reduces the embedding channel from $ C $ to $ C/4 $. 
%
%

Based on the self-similarity matrix $ {\bf A}_{\text{T}\to\text{T}} $, we transform the template feature through $ {\bf A}_{\text{T}\to\text{T}}{\bf T}^{'} $, which is added to the original feature $ {\bf T}^{'} $ as a residual term as follows:
\begin{equation}\label{self-T}
\hat{\bf T} = \text{Ins.~Norm}\left({\bf A}_{\text{T}\to\text{T}}{\bf T}^{'} + {\bf T}^{'}\right),
\end{equation}
where $ \hat{\bf T} \in \mathbb{R}^{N_T\times C} $ is the encoded template feature and $ \text{Ins.~Norm}(\cdot) $ denotes the instance normalization that jointly $ \ell_2 $-normalizes all the embeddings from an image patch, \emph{i.e.,} feature map level (${\bf T}_i \in \mathbb{R}^{C\times H \times W} $) normalization.

Thanks to the self-attention, multiple temporally diverse template features aggregate each other to generate high-quality $ \hat{\bf T} $, which is further fed to the decoder block to reinforce the search patch feature.
Besides, this encoded template representation $ \hat{\bf T} $ is also reshaped back to $ {\bf T}_\text{encoded} \in \mathbb{R}^{n \times C \times H \times W} $ for tracking model generation, \emph{e.g.,} the DCF model in Section~\ref{tracking with transformer}.

\subsection{Transformer Decoder}\label{decoder}

Transformer decoder takes the search patch feature $ {\bf S} \in \mathbb{R}^{C\times H \times W} $ as its input. 
Similar to the encoder, we first reshape this feature to $ {\bf S}^{'} \in \mathbb{R}^{N_S \times C}$, where $ N_S = H\times W$. Then, $ {\bf S}^{'} $ is fed to the self-attention block as follows:
\begin{equation}\label{self-S}
\hat{\bf S} = \text{Ins.~Norm}\left({\bf A}_{\text{S}\to\text{S}}{\bf S}^{'} + {\bf S}^{'}\right),
\end{equation}
where $ {\bf A}_{\text{S}\to\text{S}} = \text{Atten}\left(\varphi({\bf S}^{'}), \varphi({\bf S}^{'})\right)  \in \mathbb{R}^{N_S\times N_S}$ is the self-attention matrix of the search feature.

{\flushleft \bf Mask Transformation.}
Based on the search feature $ \hat{\bf S} $ in Eq.~\ref{self-S} and aforementioned encoded template feature $ \hat{\bf T} $ in Eq.~\ref{self-T}, we compute the cross-attention matrix between them via ${\bf A}_{\text{T}\to\text{S}} = \text{Atten}\left(\phi(\hat{\bf S}), \phi(\hat{\bf T})\right)  \in \mathbb{R}^{N_S\times N_T}$, where $ \phi(\cdot) $ is a $ 1\times 1 $ linear transformation block similar to $ \varphi(\cdot) $.
This cross-attention map ${\bf A}_{\text{T}\to\text{S}}$ establishes the pixel-to-pixel correspondence between frames, which supports the temporal context propagation.

In visual tracking, we are aware of the target positions in the templates.
To propagate the temporal motion priors, we construct the Gaussian-shaped masks of the template features through $ {\bf m}(y) = \text{exp}\left(-\frac{\|y - c\|^2}{2 \sigma^2}\right) $, where $ c $ is the ground-truth target position.
Similar to the feature ensemble $ \bf T $, we also concatenate these masks $ {\bf m}_{i} \in \mathbb{R}^{H\times W} $ to form the mask ensemble $ {\bf M}=\text{Concat}({\bf m}_1, \cdots, {\bf m}_n) \in \mathbb{R}^{n \times H \times W}$, which is further flattened into $ {\bf M}^{'} \in \mathbb{R}^{N_T \times 1} $.
Based on the cross attention map $ {\bf A}_{\text{T}\to\text{S}} $, we can easily propagate previous masks to the search patch via $ {\bf A}_{\text{T}\to\text{S}}{\bf M}^{'} \in \mathbb{R}^{N_S \times 1} $. 
The transformed mask is qualified to serve as the attention weight for the search feature $ \hat{\bf S} $ as follows:
\begin{equation}\label{key}
\hat{\bf S}_{\text{mask}} = \text{Ins.~Norm}\left({\bf A}_{\text{T}\to\text{S}}{\bf M}^{'} \otimes \hat{\bf S}\right),
\end{equation}
where $ \otimes $ is the broadcasting element-wise multiplication. By virtue of the spatial attention, the reinforced search feature $ \hat{\bf S}_{\text{mask}} $ better highlights the potential target area.

{\flushleft \bf Feature Transformation.} 
Except for the spatial attention, it is also feasible to propagate the context information from template feature $ \hat{\bf T} $ to the search feature $ \hat{\bf S} $. 
It is beneficial to convey target representations while the background scenes tend to change drastically in a video, which is unreasonable to temporally propagate.
As a consequence, before feature transformation, we first mask the template feature through $ \hat{\bf T}\otimes \bf{M}^{'} $ to suppress the background area.
Then, with the cross-attention matrix $ {\bf A}_{\text{T}\to\text{S}} $, the transformed feature can be computed via $ {\bf A}_{\text{T}\to\text{S}}(\hat{\bf T}\otimes {\bf M}^{'}) \in \mathbb{R}^{N_S \times C}$, which is added to $ \hat{\bf S} $ as a residual term:
\begin{equation}\label{key}
\hat{\bf S}_{\text{feat}} = \text{Ins.~Norm}\left({\bf A}_{\text{T}\to\text{S}}(\hat{\bf T}\otimes \bf{M}^{'}) + \hat{\bf S}\right).
\end{equation}
Compared with original $ \hat{\bf S} $, feature-level enhanced $ \hat{\bf S}_{\text{feat}} $ aggregates temporally diverse target representations from a series of template features $ \hat{\bf T} $ to promote itself.

Finally, we equally combine the aforementioned spatially masked feature $ \hat{\bf S}_{\text{mask}} $ and feature-level enhanced feature $ \hat{\bf S}_{\text{feat}} $, and further normalize them as follows:
\begin{equation}\label{key}
\hat{\bf S}_{\text{final}} = \text{Ins.~Norm}\left(\hat{\bf S}_{\text{feat}} + \hat{\bf S}_{\text{mask}}\right).
\end{equation}
The final output feature $ \hat{\bf S}_{\text{final}} \in \mathbb{R}^{N_S \times C}$ is reshaped back to the original size for visual tracking. 
We denote the reshaped version of $ \hat{\bf S}_{\text{final}} $ as $ {\bf S}_{\text{decoded}} \in \mathbb{R}^{C\times H \times W} $.

\begin{figure}
	\centering
	\includegraphics[width=7.3cm]{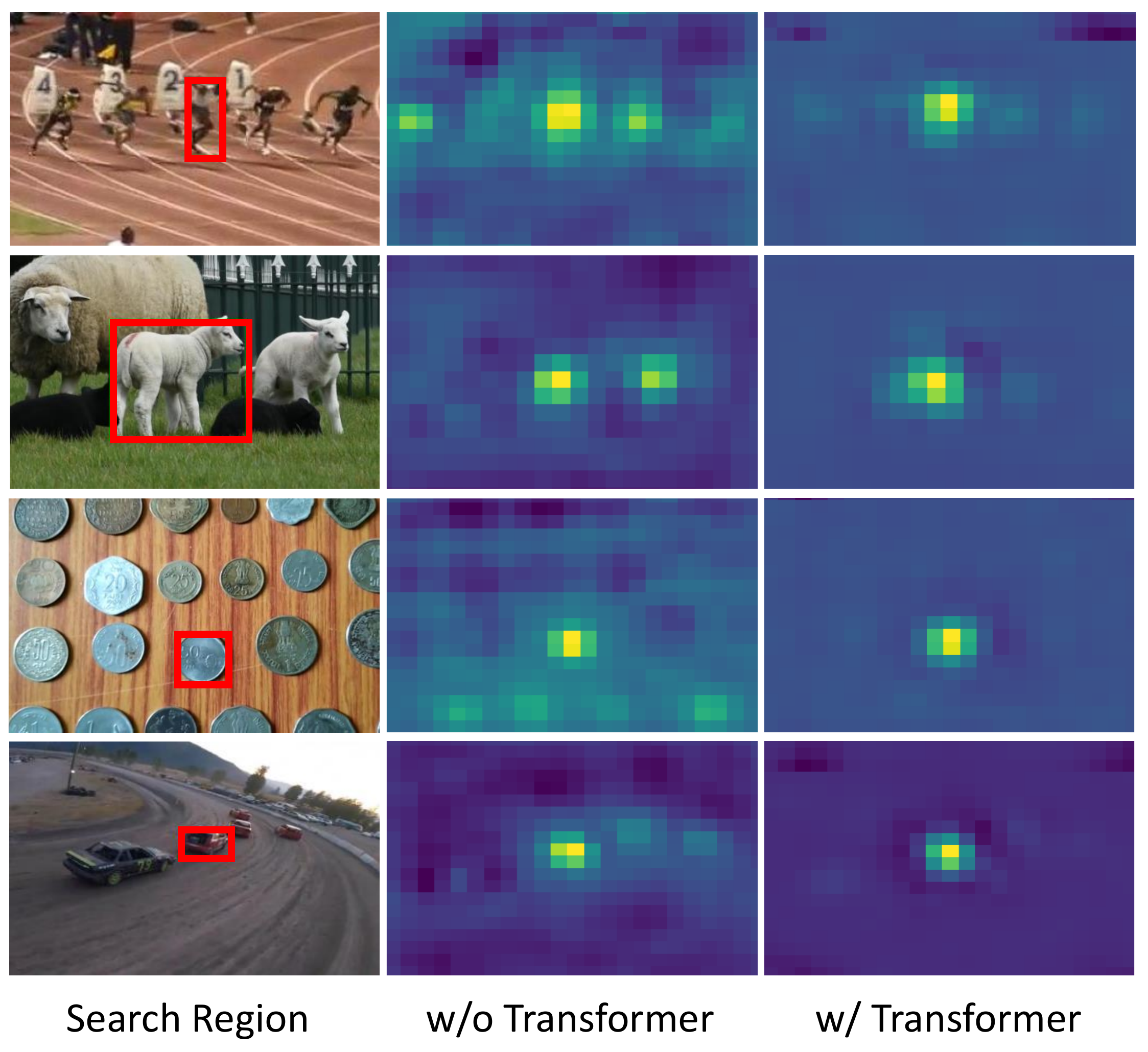}
	\caption{Tracking response maps of the DiMP baseline \cite{DiMP} without (second column) and with (third column) our designed transformer architecture. With the proposed transformer, the confidences of the distracting objects are effectively suppressed.}\label{fig:vis}
	\vspace{-0.1in}
\end{figure}

\subsection{Tracking with Transformer-enhanced Features}\label{tracking with transformer}
Transformer structure facilitates the tracking process by generating high-quality template feature $ {\bf T}_{\text{encoded}} $ and search feature $ {\bf S}_{\text{decoded}} $.
We learn the tracking model using $ {\bf T}_{\text{encoded}} $ following two popular paradigms:
\begin{itemize}[noitemsep,nolistsep]	
	\item \emph{\bf Siamese Pipeline.} In this setting, we simply crop the target feature in $ {\bf T}_{\text{encoded}} $ as the template CNN kernel to convolve with $ {\bf S}_{\text{decoded}} $ for response generation, which is identical to the cross-correlation in SiamFC \cite{SiamFC}.
	
	\item \emph{\bf DCF Pipeline.} Following the end-to-end DCF optimization in DiMP approach \cite{DiMP}, we generate a discriminative CNN kernel using $ {\bf T}_{\text{encoded}} $  to convolve with $ {\bf S}_{\text{decoded}} $ for response generation. 
\end{itemize}
After obtaining the tracking response, we utilize the classification loss proposed in DiMP \cite{DiMP} to jointly train the backbone network, our transformer, and the tracking model in an end-to-end manner. 
Please refer to \cite{DiMP} for more details.

In the online tracking process, to better exploit the temporal cues and adapt to the target appearance changes, we dynamically update the template ensemble $ \bf T $.
To be specific, we drop the oldest template in $ \bf T $ and add the current collected template feature to $ \bf T $ every 5 frames.
The feature ensemble maintains a maximal size of 20 templates.
Once the template ensemble $ \bf T $ is updated, we compute the new encoded feature $ {\bf T}_{\text{encoded}} $ via our transformer encoder.
While the transformer encoder is sparsely utilized (\emph{i.e.,} every 5 frames), the transformer decoder is leveraged in each frame, which generates per-frame $ {\bf S}_{\text{decoded}} $ by propagating the representations and attention cues from previous templates to the current search patch.

It is widely recognized that DCF formulation in DiMP \cite{DiMP} is superior to the simple cross-correlation in Siamese trackers \cite{SiamFC,siamrpn++}.
Nevertheless, in the experiments, we show that with the help of our transformer architecture, a classic Siamese pipeline is able to perform against the recent DiMP.
Meanwhile, with our transformer, the DiMP tracker acquires further performance improvements.
As shown in Figure~\ref{fig:vis}, even though the strong baseline DiMP \cite{DiMP} already shows impressive distractor discrimination capability, our designed transformer further assists it to restrain the background confidence for robust tracking.

\section{Experiments}

\subsection{Implementation Details}

Based on the Siamese matching and DiMP based tracking frameworks, in the following experiments, we denote our \underline{Tr}ansformer-assisted trackers as {\bf TrSiam} and {\bf TrDiMP}, respectively.
In these two versions, the backbone model is ResNet-50 \cite{ResNet} for feature extraction.
Before the encoder and decoder, we additionally add one convolutional layer (3$ \times $3 Conv + BN) to reduce the backbone feature channel from 1024 to 512.
The input template and search patches are 6 times of the target size and further resized to 352$ \times $352.
The temperature $ \tau $ in Eq.~\ref{EQ3} is set to 1/30. The parameter sigma $ \sigma $ in the feature mask is set to 0.1.
Similar to the previous works \cite{ATOM,DiMP,PrDiMP,KYS}, we utilize the training splits of LaSOT \cite{LaSOT}, TrackingNet \cite{2018trackingnet}, GOT-10k \cite{GOT10k}, and COCO \cite{COCO} for offline training.
The proposed transformer network is jointly trained with the original tracking parts (\emph{e.g.,} tracking optimization model \cite{DiMP} and IoUNet \cite{PrDiMP}) in an end-to-end manner. 
Our framework is trained for 50 epochs with 1500 iterations per epoch and 36 image pairs per batch.
The ADAM optimizer \cite{ADAM} is employed with an initial learning rate of 0.01 and a decay factor of 0.2 for every 15 epochs.

In the online tracking stage, the main difference between TrSiam and TrDiMP lies in the tracking model generation manner.
After predicting the response map for target localization, they all adopt the recent probabilistic IoUNet \cite{PrDiMP} for target scale estimation.
Our trackers are implemented in Python using PyTorch. 
TrSiam and TrDiMP operate about 35 and 26 frames per second (FPS) on a single Nvidia GTX 1080Ti GPU, respectively.
%
%
%

\subsection{Ablation Study}
\begin{figure}[t]
	\centering
	\includegraphics[width=4.12cm]{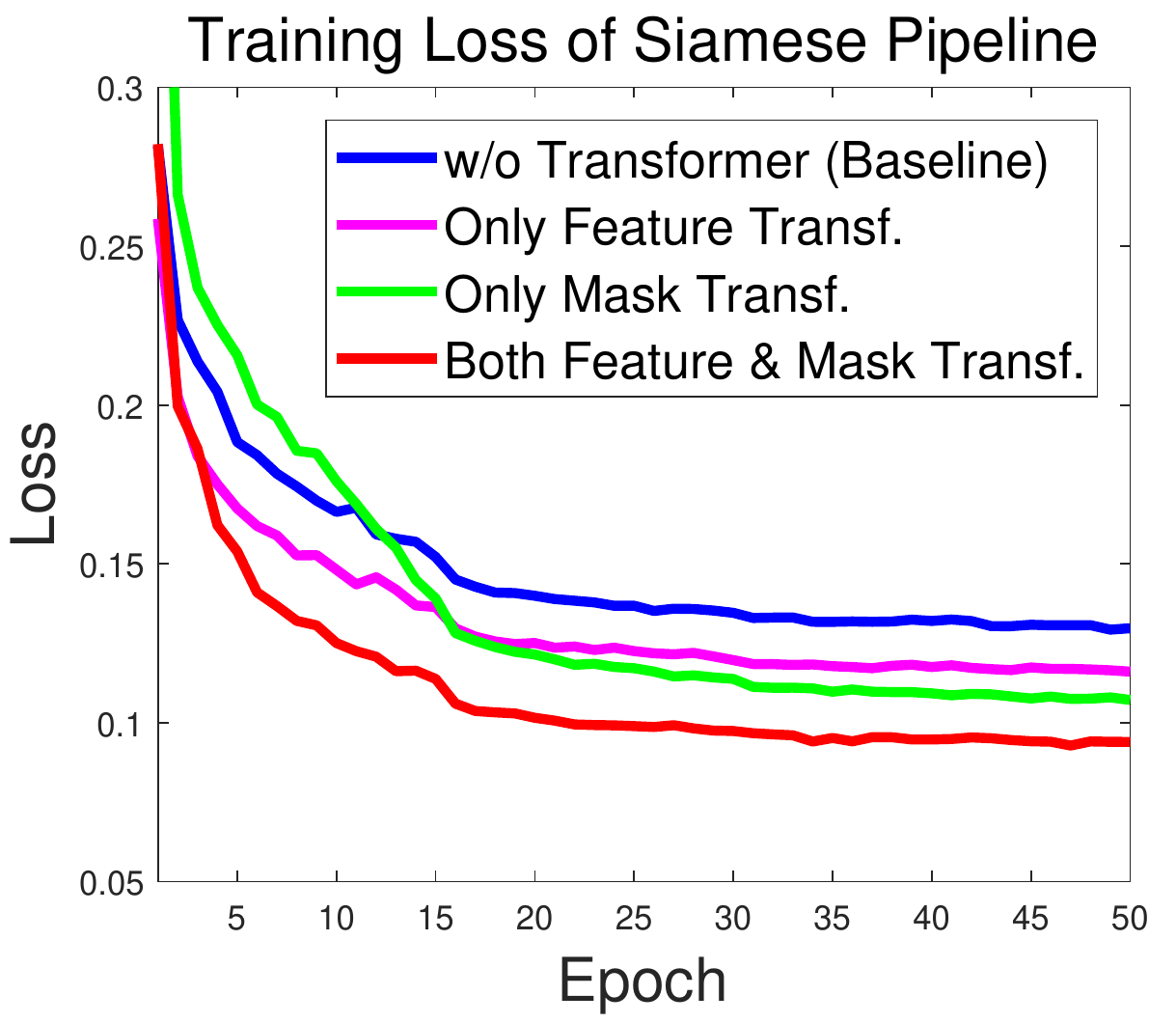}
	\includegraphics[width=4.12cm]{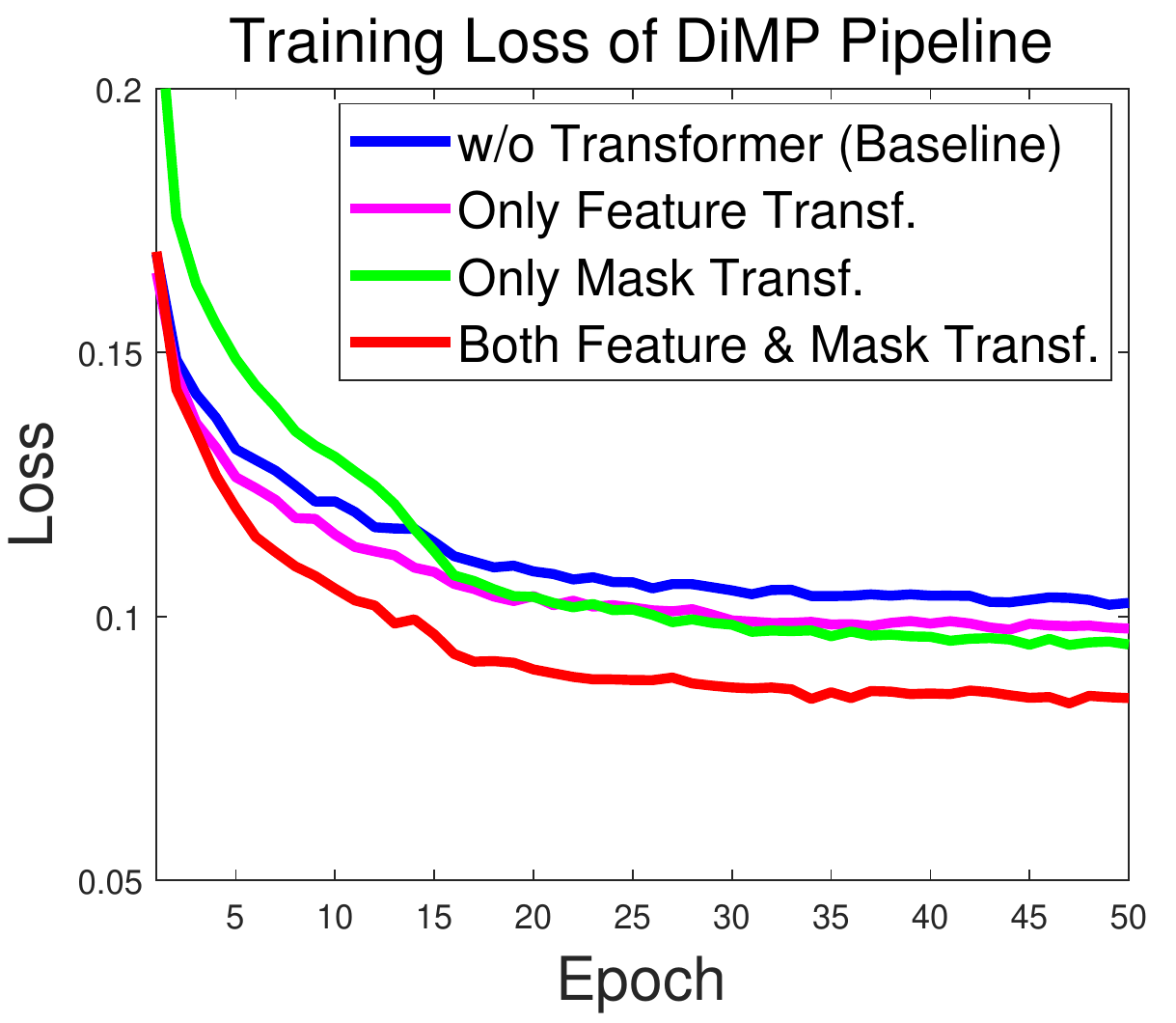}
	\caption{Training loss plots of the Siamese pipeline (left) and DCF pipeline (right). By combining both feature and mask transformations, our approach significantly reduces the training losses.} 
	\label{fig:loss plot} 
	\vspace{-0.0in}
\end{figure}

To verify the effectiveness of our designed transformer structure, we choose the GOT-10k test set \cite{GOT10k} with 180 videos to validate our TrSiam and TrDiMP methods\footnote{With the probabilistic IoUNet \cite{PrDiMP} and a larger search area, our baseline performance is better than the standard DiMP \cite{DiMP}. Note that all the experiments (Figure~\ref{fig:loss plot} and Table~\ref{table:ablation study_both}) are based on the same baseline for fair.}.
%
%
GOT-10k hides the ground-truth labels of the test set to avoid the overly hyper-parameter fine-tuning.
It is worth mentioning that there is no overlap in object classes between the train and test sets of GOT-10k, which also verifies the generalization of our trackers to unseen object classes.
%

In Table~\ref{table:ablation study_both}, based on the Siamese and DiMP baselines, we validate each component in our transformer:

{\noindent \bf Transformer Encoder.} First, without any decoder block, we merely utilize encoder to promote the feature fusion of multiple templates, which slightly improves two baselines.
%

{\noindent \bf Transformer Decoder.}
Our decoder consists of feature and mask transformations, and we independently verify them:
{\noindent (1)} \emph{Feature Propagation.} 
With the feature transformation, as shown in Table~\ref{table:ablation study_both}, the Siamese pipeline obtains a notable performance gain of 4.3\% in AO and the strong DiMP baseline still acquires an improvement of 1.4\% in AO on the GOT-10k test set.
From the training perspective, we can observe that this block effectively reduces the losses of two baselines as shown in Figure~\ref{fig:loss plot}.

{\noindent (2)} \emph{Mask Propagation.} This mechanism propagates temporally collected spatial attentions to highlight the target area.
Similar to the feature transformation, our mask transformation alone also steadily improves the tracking performance (Table~\ref{table:ablation study_both}) and consistently reduces the training errors of both two pipelines (Figure~\ref{fig:loss plot}).

\setlength{\tabcolsep}{2pt}
\begin{table}[t]
	\scriptsize
	\begin{center}
		\caption{Ablative experiments of our transformer for the Siamese and DiMP pipelines \emph{i.e.,} TrSiam and TrDiMP trackers. The performance is evaluated on the GOT-10k test set \cite{GOT10k} in terms of average overlap (AO).} \label{table:ablation study_both}	
		\begin{tabular*}{8.4 cm} {@{\extracolsep{\fill}}lc|cc}
			\hline
			&Different Tracking Variations &Siamese (AO) &DiMP (AO) ~\\
			\hline
			&Baseline Performance   & 62.0 &66.7 ~ \\
			&Only Encoder (w/o Any Decoder)   &${\text{63.8}}_{1.8\%\uparrow} $ &${\text{67.3}}_{0.6\%\uparrow} $ \\
			&Encoder +  Decoder (Only Feature Transf.)  &${\text{66.3}}_{4.3\%\uparrow} $ &${\text{68.1}}_{1.4\%\uparrow} $ ~\\
			&Encoder +  Decoder (Only Mask Transf.) &${\text{67.1}}_{5.1\%\uparrow} $ &${\text{67.8}}_{1.1\%\uparrow} $ ~\\
			&Encoder + Decoder (Feature \& Mask Transf.)  &${\text{\bf 67.3}}_{\bf 5.3\%\uparrow} $ &${\text{\bf 68.8}}_{\bf 2.1\%\uparrow} $~ \\
			\hline
		\end{tabular*}
	\end{center}
	\vspace{-0.1in}
\end{table}

\setlength{\tabcolsep}{2pt}
\begin{table}[t]
	\scriptsize
	\begin{center}
		\caption{Ablative study of our transformer architecture. The baseline tracker is TrSiam. The evaluation metric is average overlap (AO) score on the GOT-10k test set.} \label{table: design}	
		\begin{tabular*}{8.4 cm} {@{\extracolsep{\fill}}lc|cc|cc|ccc}
			\hline
			& Baseline & \multicolumn{2}{c|}{Weight-sharing} &\multicolumn{2}{c|}{Feed-forward} &\multicolumn{3}{c}{Head Number}  \\
			
			& &w/o & w/ &w/o &w/ &1  &2 &4\\
			\hline
			AO (\%) &62.0 &63.4 &{\bf 67.3} &{\bf 67.3} &{67.0} &67.3 &67.2 &{\bf 67.6}\\
			Speed (FPS) &40 &35 &35 &35  &22 &35 &31 &25\\
			\hline
		\end{tabular*}
	\end{center}
	\vspace{-0.2in}
\end{table}

\setlength{\tabcolsep}{2pt}
\begin{table*}[t]
	\scriptsize
	\begin{center}
		\caption{Comparison with state-of-the-art trackers on the TrackingNet test set \cite{2018trackingnet} in terms of precision (Prec.), normalized precision (N. Prec.), and success (AUC score). Our TrDiMP and TrSiam exhibit promising results.} \label{table:trackingnet}	
		\begin{tabular*}{17.6 cm} {@{\extracolsep{\fill}}lcccccccccccccccc}
			\hline
			~  &SiamFC &MDNet &SPM &C-RPN &SiamRPN++ &ATOM &DiMP-50 &SiamFC++ &D3S &Retain-MAML &PrDiMP-50 &DCFST &KYS &Siam-RCNN &{\bf TrSiam} &{\bf TrDiMP}\\
			~  &\cite{SiamFC}  &\cite{MDNet} &\cite{SPM} &\cite{CRPN} &\cite{siamrpn++} &\cite{ATOM} &\cite{DiMP} &\cite{SiamFC++} &\cite{D3S} &\cite{MAML} &\cite{PrDiMP} &\cite{DCFST} &\cite{KYS} &\cite{SiamRCNN} & & \\
			\hline
			~Prec. (\%) &53.3  &56.5  &66.1 &61.9 &69.4 &64.8 &68.7 &70.5 &66.4 &- &70.4 &70.0 &68.8 &{\bf \color{red} 80.0} &{ 72.7} &{\bf \color{blue} 73.1}\\
			
			~N. Prec. (\%)  &66.3  &70.5 &77.8 &74.6 &80.0 &77.1 &80.1 &80.0 &76.8 &82.2 &81.6 &80.9 &80.0 &{\bf \color{red} 85.4} &{ 82.9} &{\bf \color{blue} 83.3}\\
			
			~Success (\%)  &57.1  &60.6 &71.2 &66.9 &73.3 &70.3 &74.0 &75.4 &72.8 &75.7 &75.8 &75.2 &74.0 &{\bf \color{red} 81.2} &{ 78.1} &{\bf \color{blue} 78.4}\\
			\hline
		\end{tabular*}
	\end{center}
	\vspace{-0.15in}
\end{table*}

\setlength{\tabcolsep}{2pt}
\begin{table*}[t]
	\scriptsize
	\begin{center}
		\caption{Comparison results on the GOT-10k test set \cite{GOT10k} in terms of average overlap (AO), and success rates (SR) at overlap thresholds 0.5 and 0.75. We show the tracking results without (w/o) and with (w/) additional training data (LTC: LaSOT, TrackingNet, and COCO).} \label{table:got10k}	
		\begin{tabular*}{17.5 cm} {@{\extracolsep{\fill}}lcccccccccccccccc}
			\hline
			~   &SiamFC &SiamFCv2 &SiamRPN &SPM &ATOM &DiMP-50 &SiamFC++ &D3S &PrDiMP-50 &DCFST &KYS &Siam-RCNN &\multicolumn{2}{c}{\bf TrSiam} &\multicolumn{2}{c}{\bf TrDiMP}\\
			~   &\cite{SiamFC} &\cite{CFNet} &\cite{SiamRPN} &\cite{SPM} &\cite{ATOM} &\cite{DiMP} &\cite{SiamFC++} &\cite{D3S} &\cite{PrDiMP} &\cite{DCFST} &\cite{KYS} &\cite{SiamRCNN} &w/o LTC  &w/ LTC &w/o LTC &w/ LTC\\
			\hline
			
			$\text{SR}_{0.5} $(\%)   &35.3 &40.4 &54.9 &59.3 &63.4 &71.7 &69.5 &67.6 &73.8 &75.3 &75.1 &- &{\bf \color{blue} 76.6} &{\bf \textcolor[rgb]{0.5,0.5,0.5}{78.7}} &{\bf \color{red} 77.7} &{\bf \textcolor[rgb]{0.5,0.5,0.5}{80.5}}\\
			
			$\text{SR}_{0.75} $(\%)   &9.8 &14.4 &25.3 &35.9 &40.2 &49.2 &47.9 &46.2 &54.3 &49.8 &51.5 &- &{\bf \color{blue} 57.1} &{\bf \textcolor[rgb]{0.5,0.5,0.5}{58.6}} &{\bf \color{red} 58.3} &{\bf \textcolor[rgb]{0.5,0.5,0.5}{59.7}}\\
			
			AO (\%) &34.8 &37.4 &46.3 &51.3 &55.6 &61.1 &59.5 &59.7 &63.4 &63.8 &63.6 &64.9 &{\bf \color{blue} 66.0} &{\bf \textcolor[rgb]{0.5,0.5,0.5}{67.3}} &{\bf \color{red} 67.1} &{\bf \textcolor[rgb]{0.5,0.5,0.5}{68.8}}\\
			\hline
		\end{tabular*}
	\end{center}
	\vspace{-0.15in}
\end{table*}

{\noindent \bf Complete Transformer.} 
With the complete transformer, as shown in Table~\ref{table:ablation study_both}, the Siamese and DiMP baselines obtain notable performance gains of 5.3\% and 2.1\% in AO, respectively.
The transformer also significantly reduces their training losses (Figure~\ref{fig:loss plot}).
It is worth mentioning that DiMP already achieves outstanding results while our approach consistently improves such a strong baseline.
%
%
With our transformer, the performance gap between Siamese and DiMP baselines has been largely narrowed (from 4.7\% to 1.5\% in AO), which reveals the strong tracking potential of a simple pipeline by adequately exploring the temporal information.
%

{\noindent \bf Structure Modifications.} Finally, we discuss some architecture details of our transformer:
{(1)} \emph{Shared-weight Self-attention}. Since our transformer is separated into two parallel Siamese tracking braches, the performance obviously drops without the weight-sharing mechanism as shown in Table~\ref{table: design}.
Due to this weight-sharing design, we also do not stack multiple encoder/decoder layers like the classic transformer \cite{Transformer}, which will divide the template and search representations into different feature subspaces.
{(2)} \emph{Feed-forward Network}. Feed-forward network is a basic block in the classic transformer \cite{Transformer}, which consists of two heavyweight fully-connected layers.
In the tracking scenario, we observe that this block potentially causes the overfitting issue due to its overmany parameters, which does not bring performance gains and hurts the efficiency.
{(3)}  \emph{Head Number}. Classic transformer adopts multi-head attentions (\emph{e.g.,} 8 heads) to learn diverse representations \cite{Transformer}. 
In the experiments, we observe that increasing the head number slightly improves the accuracy but hinders the tracking efficiency from real-time.
We thus choose the single-head attention to achieve a good balance of performance and efficiency.
%

\subsection{State-of-the-art Comparisons}

We compare our proposed TrSiam and TrDiMP trackers with the recent state-of-the-art trackers on seven tracking benchmarks including TrackingNet \cite{2018trackingnet}, GOT-10k \cite{GOT10k}, LaSOT \cite{LaSOT}, VOT2018 \cite{VOT2018}, Need for Speed \cite{NFSdataset}, UAV123 \cite{UAV123}, and OTB-2015 \cite{OTB-2015}.

{\noindent \bf TrackingNet \cite{2018trackingnet}.} TrackingNet is a recently released large-scale benchmark. We evaluate our methods on the test set of TrackingNet, which consists of 511 videos. 
In this benchmark, we compare our approaches with the state-of-the-art trackers such as DiMP-50 \cite{DiMP}, D3S \cite{D3S}, SiamFC++ \cite{SiamFC++}, Retain-MAML \cite{MAML}, DCFST \cite{DCFST}, PrDiMP-50 \cite{PrDiMP}, KYS \cite{KYS}, and Siam-RCNN \cite{SiamRCNN}.
As shown in Table~\ref{table:trackingnet}, the proposed TrDiMP achieves a normalized precision score of 83.3\% and a success score of 78.4\%, surpassing previous state-of-the-art trackers such as PrDiMP-50 and KYS.
Note that PrDiMP and KYS improve the DiMP tracker via probabilistic regression and tracking scene exploration, representing the current leading algorithms on several datasets.
With our designed transformer, the simple Siamese matching baseline (\emph{i.e.,} TrSiam) also shows outstanding performance with a normalized precision score of 82.9\% and a success score of 78.1\%. 

\begin{figure}[t]
	\centering
	\includegraphics[width=4.12cm]{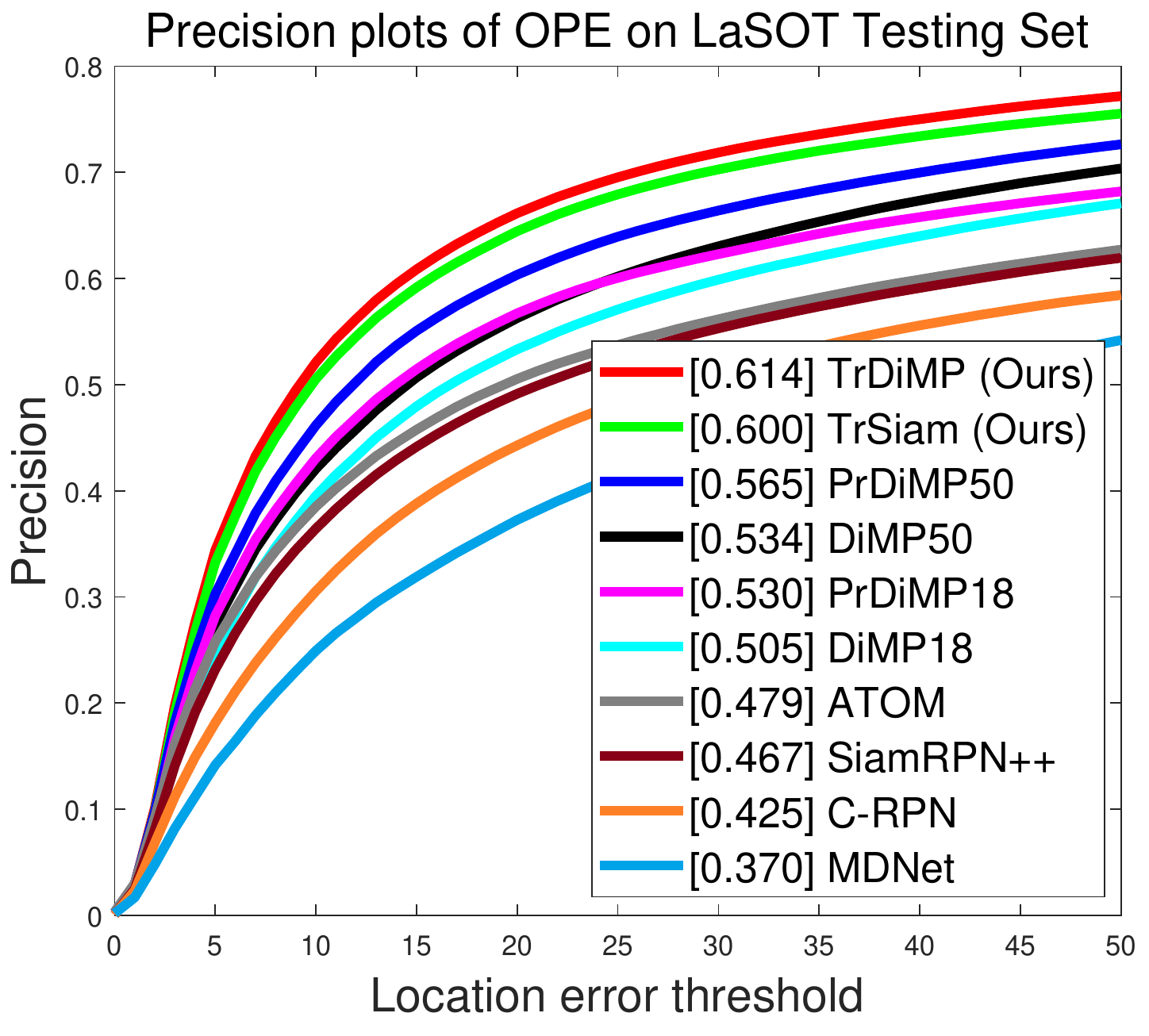}
	\includegraphics[width=4.12cm]{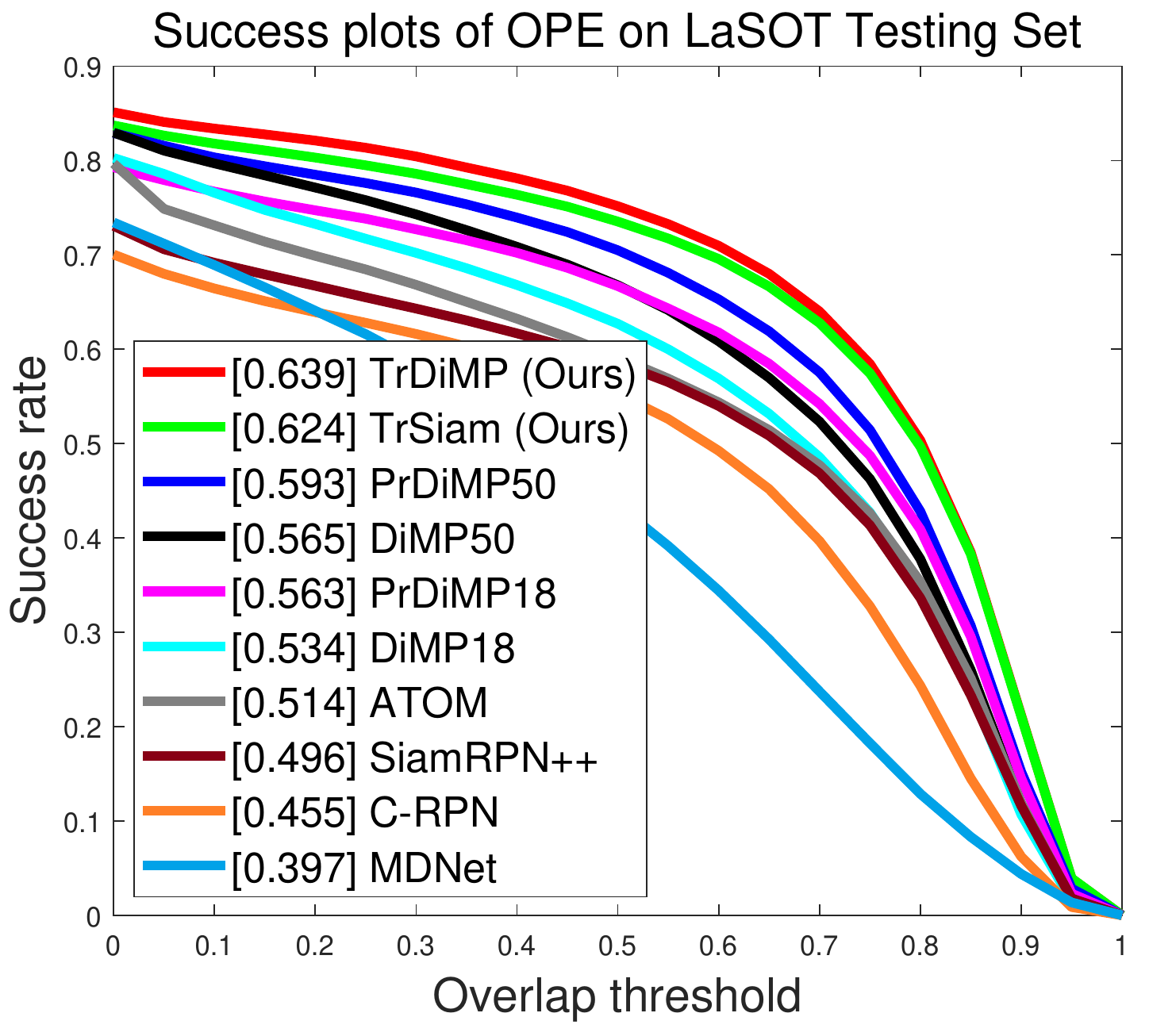}
	\caption{Precision and success plots on the LaSOT  test set \cite{LaSOT}. In the legend, the distance precision (DP) and area-under-curve (AUC) are reported in the left and right figures, respectively.}
	\label{fig:lasot} 
	\vspace{-0.18in}
\end{figure}

\setlength{\tabcolsep}{2pt}
\begin{table*}[t]
	\scriptsize
	\begin{center}
		\caption{State-of-the-art comparison on the NfS \cite{NFSdataset}, UAV123 \cite{UAV123}, and OTB-2015 \cite{OTB-2015} datasets in terms of AUC score. Both our TrDiMP and TrSiam exhibit outstanding results on all benchmarks with competitive efficiency.} \label{table:uav_nfs}	
		\vspace{+0.06in}
		\begin{tabular*}{17.4 cm} {@{\extracolsep{\fill}}lcccccccccccccccc}
			\hline
			~ & KCF &SiamFC &CFNet &MDNet &C-COT &ECO &ATOM &UPDT &SiamRPN++ &DiMP-50 &SiamR-CNN &PrDiMP-50 &DCFST &KYS &{\bf TrSiam} &{\bf TrDiMP}\\
			
			~ &\cite{KCF} &\cite{SiamFC} &\cite{CFNet} &\cite{MDNet} &\cite{C-COT} &\cite{ECO} &\cite{ATOM} &\cite{UPDT} &\cite{siamrpn++}  &\cite{DiMP} &\cite{SiamRCNN} &\cite{PrDiMP} &\cite{DCFST} &\cite{KYS} & &\\
			\hline
			
			NfS \cite{NFSdataset}  &21.7 &- &- &42.9 &48.8 &46.6 &58.4 &53.7 &50.2 &62.0 &63.9 &63.5 &{ 64.1} &63.5 &{\bf \color{blue} 65.8} &{\bf \color{red} 66.5}\\
			
			UAV123 \cite{UAV123}  &33.1 &49.8 &43.6 &52.8 &51.3 &52.2 &64.2 &54.5 &61.3 &65.3 &64.9 &{\bf \color{red} 68.0} &- &- &{ 67.4} &{\bf \color{blue} 67.5}\\
			
			OTB-2015 \cite{OTB-2015} &47.5 &58.2 &56.8 &67.8 &68.2 &69.1 &66.9 &70.2 &69.6 &68.4 &70.1 &69.6 &{\bf \color{blue} 70.9} &69.5 &{ 70.8} &{\bf \color{red} 71.1}\\
			\hline
			Speed (FPS) &{\bf \color{red} 270} &{\color{blue} \bf 86} &75 &1 &0.3 &8 &35 &$ < $1 &30 &35 &4.7 &30 &25 &20 &35 &26\\
			\hline
		\end{tabular*}
	\end{center}
	\vspace{-0.2in}
\end{table*}

{\noindent \bf GOT-10k \cite{GOT10k}.} GOT-10k is a large-scale dataset including more than 10,000 videos. We test our methods on the test set of GOT-10k with 180 sequences. 
The main characteristic of GOT-10k is that the test set does not have overlap in object classes with the train set, which is designed to assess the generalization of the visual tracker.
%
Following the test protocol of GOT-10k, we further train our trackers with only the GOT-10k training set.
As shown in Table \ref{table:got10k}, in a fair comparison scenario (\emph{i.e.,} without additional training data), both our TrDiMP and TrSiam still outperform other top-performing trackers such as SiamR-CNN \cite{SiamRCNN}, DCFST \cite{DCFST}, and KYS \cite{KYS}, verifying the strong generalization of our methods to unseen objects.
%

{\noindent \bf LaSOT \cite{LaSOT}.} LaSOT is a recent large-scale tracking benchmark consisting of 1200 videos. The average video length of this benchmark is about 2500 frames, which is more challenging than the previous short-term tracking datasets. 
Therefore, how to cope with the drastic target appearance changes using temporal context is vital in this dataset. 
We evaluate our approaches on the LaSOT test set with 280 videos. 
The precision and success plots of the state-of-the-art methods are shown in Figure \ref{fig:lasot}, where the recently proposed C-RPN \cite{CRPN}, SiamRPN++ \cite{siamrpn++}, ATOM \cite{ATOM}, DiMP-50 \cite{DiMP}, and PrDiMP-50 \cite{PrDiMP} are included for comparison.
Our TrSiam and TrDiMP outperform aforementioned methods by a considerable margin.
%
%
To the best of our knowledge, SiamR-CNN \cite{SiamRCNN} achieves the current best result on the LaSOT.
Overall, our TrDiMP (63.9\% AUC and 26 FPS) exhibits very competitive performance
and efficiency in comparison with SiamR-CNN (64.8\% AUC and 4.7 FPS).

\begin{figure}[t]
	\centering
	\includegraphics[width=8.0cm]{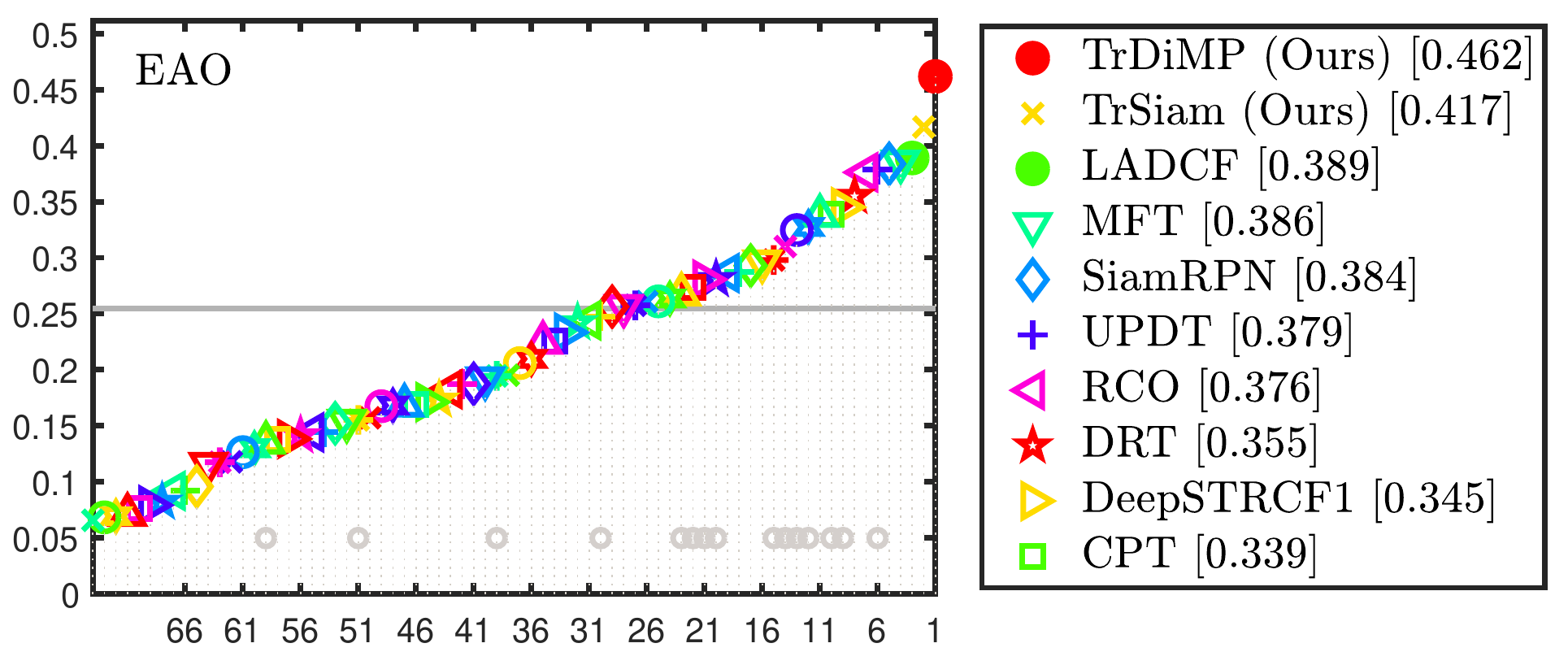}
	\caption{Expected average overlap (EAO) graph with trackers ranked from right to left. Our TrDiMP and TrSiam trackers outperform all the participant trackers on the VOT2018 \cite{VOT2018}.}\label{fig:vot} 
	\vspace{-0.1in}
\end{figure}

{\noindent \bf VOT2018 \cite{VOT2018}.} VOT2018 benchmark contains 60 challenging videos. 
The performance on this dataset is evaluated using the expected
average overlap (EAO), which takes both accuracy (average overlap over successful frames) and robustness (failure rate) into account. 
%
%
As shown in Figure~\ref{fig:vot}, our TrSiam and TrDiMP clearly outperform all the participant trackers on the VOT2018.

In Table~\ref{table:vot}, we further show the accuracy, robustness, and EAO scores of the recent top-performing trackers including SiamRPN++ \cite{siamrpn++}, DiMP-50 \cite{DiMP}, PrDiMP-50 \cite{PrDiMP}, Retain-MAML \cite{MAML}, KYS \cite{KYS}, and D3S \cite{D3S}.
Compared with these recently proposed approaches, our TrDiMP approach still exhibits satisfactory results.
%
%
Among all the compared trackers, only D3S slightly outperforms our TrDiMP, which is trained using additional data with segmentation annotations for accurate mask prediction.

{\noindent \bf NfS \cite{NFSdataset}.} 
NfS dataset contains 100 challenging videos with fast-moving objects.
We evaluate our TrSiam and TrDiMP on the 30 FPS version of NfS.
The AUC scores of comparison approaches are shown in Table~\ref{table:uav_nfs}. 
Our approaches set new state-of-the-art records on this benchmark. 
The proposed TrDiMP surpasses previous top-performing trackers such as DCFST \cite{DCFST} and SiamR-CNN \cite{SiamRCNN}.
Note that the recent SimR-CNN utilizes a powerful ResNet-101 for object re-detection. 
Our simple TrSiam, without sophisticated models or online optimization techniques, still outperforms existing methods and operates in real-time.

{\noindent \bf UAV123 \cite{UAV123}.} 
This benchmark includes 123 aerial videos collected by the low-attitude UAV platform.
The proposed trackers also achieve promising results in comparison to the recent remarkable approaches in Table~\ref{table:uav_nfs}. 
Specifically, our TrDiMP performs on par with PrDiMP-50 \cite{PrDiMP}, which represents the current best algorithm on this benchmark. 

\setlength{\tabcolsep}{2pt}
\begin{table}[t]
	\scriptsize
	\begin{center}
		\caption{Comparison with recent state-of-the-art trackers on the VOT2018 \cite{VOT2018} in terms of accuracy (A), robustness (R), and expected average overlap (EAO).} \label{table:vot}	
		\begin{tabular*}{8.4 cm} {@{\extracolsep{\fill}}lccccccc}
			\hline
			~   & SiamRPN &DiMP-50 &PrDiMP-50 &Retain- &KYS &D3S  &{\bf TrDiMP} \\
			~   &++ \cite{siamrpn++} &\cite{DiMP} &\cite{PrDiMP} &MAML \cite{MAML} &\cite{KYS} &\cite{D3S}  & \\
			\hline
			~A ($ \uparrow $)       &0.600  &0.597 &{\bf \color{blue} 0.618} &0.604 &0.609 &{\bf \color{red}0.640} &0.600 \\
			~R ($ \downarrow $)       &0.234  &0.153 &0.165 &0.159 &{\bf \color{blue} 0.143} &0.150 &{\bf \color{red} 0.141} \\
			~EAO ($ \uparrow $)     &0.414  &0.440 &0.442 &0.452 &{\bf \color{blue} 0.462} &{\bf \color{red} 0.489} &{\bf \color{blue} 0.462} \\
			\hline
		\end{tabular*}
	\end{center}
	\vspace{-0.2in}
\end{table}

{\noindent \bf OTB-2015 \cite{OTB-2015}.} 
OTB-2015 is a popular tracking benchmark with 100 challenging videos.
%
%
As shown in Table~\ref{table:uav_nfs}, on this dataset, our TrDiMP achieves an AUC score of 71.1\%, surpassing the recently proposed SiamRPN++ \cite{siamrpn++}, PrDiMP-50 \cite{PrDiMP}, SiamR-CNN \cite{SiamRCNN}, and KYS \cite{KYS}. 
With the proposed transformer, our Siamese matching based TrSiam also performs favorably against existing state-of-the-art approaches with an AUC score of 70.8\%.

\section{Conclusion}
In this work, we introduce the transformer structure to the tracking frameworks, which bridges the isolated frames in the video flow and conveys the rich temporal cues across frames.
We show that by carefully modifying the classic transformer architecture, it favorably suits the tracking scenario.
With the proposed transformer, two popular trackers gain consistent performance improvements and set several new state-of-the-art records on prevalent tracking datasets.
To our best knowledge, this is the first attempt to exploit the transformer in the tracking community, which preliminarily unveils the tracking potential hidden in the frame-wise relationship.
In the future, we intend to further explore the rich temporal information among individual video frames.

\footnotesize {\flushleft \bf Acknowledgements}. This work was supported in part by the National Natural Science Foundation of China under Contract 61836011, 61822208, and 61836006, and in part by the Youth Innovation Promotion Association CAS under Grant 2018497.

{\small
\bibliographystyle{ieee_fullname}
\bibliography{AgeGender}
}

\newpage

\appendix
\renewcommand{\appendixname}{Appendix~\Alph{section}}

\normalsize

\section{Ablation Study}

\subsection{Hyper-parameters}

In the online tracking stage, the only involved hyper-parameters are the template sampling interval as well as the template ensemble size.
As shown in Table~\ref{table:interval}, we observe that sampling the template every 5 frames shows promising results.
This sparse update mechanism is also widely adopted in many previous trackers such as ECO \cite{ECO} and ATOM \cite{ATOM}.
Besides, increasing the memory size (\emph{i.e.,} the total sample number in the template ensemble $ \bf T $) also steadily improves the performance.
To achieve a good balance of performance and efficiency, we choose the maximum ensemble size of 20.

As for other tracking-related hyper-parameters, we follow our baseline approach DiMP \cite{DiMP} without modification.
More details can be found in the source code.

\subsection{Improvements upon Baselines}
In Table~\ref{table:siam_ablation} and \ref{table:dimp_ablation}, we compare our TrSiam and TrDiMP with their corresponding baselines on seven tracking benchmarks.
As shown in Table~\ref{table:siam_ablation}, our designed transformer consistently improves the Siamese baseline on seven tracking datasets.
For example, our TrSiam approach outperforms its baseline by 5.3\%, 4.7\%, 3.3\%, and 3.0\% in terms of AUC score on the challenging GOT-10k, NfS, LaSOT, and TrackingNet datasets, respectively.
On the OTB-2015 dataset, our approach still improves the baseline by 1.6\%.
The OTB-2015 dataset is known to be highly saturated over recent years. 
Note that our Siamese baseline already achieves a high performance level of 69.2\% AUC on the OTB-2015. 
Thus, it is relatively harder to obtain a significant performance gain on this benchmark.

In Table~\ref{table:dimp_ablation}, we further exhibit the comparison results between our transformer-assisted TrDiMP and its baseline DiMP \cite{DiMP}.
It is worth mentioning that the DiMP approach already introduces a memory mechanism to incrementally update the tracking model and explores the temporal information to some extent.  
Besides, our baseline includes the recent probabilistic IoUNet \cite{PrDiMP} for accurate target scale estimation and adopts a larger search area (6 times of the target object) for tracking (\emph{i.e.,} the superDiMP setting\footnote{\url{https://github.com/visionml/pytracking/tree/master/ltr}}), which significantly outperforms the standard DiMP approach presented in \cite{DiMP} .
It is well recognized that improving a strong baseline is much more challenging.
Although our baseline achieves outstanding results on various tracking benchmarks, our proposed transformer consistently improves it on \emph{all} datasets.

\setlength{\tabcolsep}{2pt}
\begin{table}[t]
	\scriptsize
	\begin{center}
		\caption{Ablation experiments on the template sampling interval and template ensemble size. The testing approach is our TrSiam. The performance is evaluated on the GOT-10k test set \cite{GOT10k} and NfS \cite{NFSdataset} in terms of AUC score.} \label{table:interval}	
		\vspace{+0.05in}
		\begin{tabular*}{8.3 cm} {@{\extracolsep{\fill}}l|cccc|cccc|cccc}
			\hline
			~Interval &1 &1 &1 &1 &5 &5 &5 &5 & 10 &10 &10 &10 \\
			~Ens. Size &1 &10 &20 &30 &1 &10 &20 &30 &1 &10 &20 &30  \\
			\hline
			~GOT-10k &63.8 &65.5 &66.1 &66.9 &63.6 &66.5 &67.3 &{\bf 67.6} &63.3 &65.5 &65.4 &65.6\\
			~NfS &62.2 &63.9 &64.5 &64.4 &62.1 &64.4 &{\bf 65.8} &{\bf 65.8} &62.2 &64.9 &65.3 &65.2\\
			\hline
			~FPS &36 &32 &28 &22 &40 &38 &35 &31 &40 &38 &36 &33 \\
			\hline
		\end{tabular*}
	\end{center}
	\vspace{-0.05in}
\end{table}

\setlength{\tabcolsep}{2pt}
\begin{table}[t]
	\scriptsize
	\begin{center}
		\caption{Comparison results of the Siamese pipeline between without and with our transformer on 7 tracking benchmarks. We compute the relative gain in the VOT2018, while in the rest datasets, we exhibit the absolute gain.} \label{table:siam_ablation}	
		\vspace{+0.05in} 
		\begin{tabular*}{8.2 cm} {@{\extracolsep{\fill}}lccc}
			\hline
			~Dataset &Siamese Baseline &{\bf TrSiam (Ours)} &$\Delta$\\
			\hline
			~Need for Speed \cite{NFSdataset} (AUC) &61.1 &\bf 65.8  &\bf 4.7\%$ \uparrow $ \\
			~OTB-2015 \cite{OTB-2015} (AUC)  &69.2 &\bf 70.8   &\bf 1.6\%$ \uparrow $\\
			~UAV123 \cite{UAV123} (AUC)  &65.6 &\bf 67.4   &\bf 1.8\%$ \uparrow $\\
			~LaSOT \cite{LaSOT} (AUC)  &59.1 &\bf 62.4  &\bf 3.3\%$ \uparrow $\\
			~GOT-10k \cite{GOT10k} (AO) &62.0 &\bf 67.3  &\bf 5.3\%$ \uparrow $\\
			~TrackingNet \cite{2018trackingnet} (Success) &75.1 &\bf 78.1  &\bf 3.0\%$ \uparrow $\\
			~VOT2018 \cite{VOT2018} (EAO)  &0.389 &\bf 0.417  &\bf 7.2\%$ \uparrow $ \\
			\hline
			~Tracking Speed (FPS) &\bf 40  &35 & 5 FPS $ \downarrow $ \\
			\hline
		\end{tabular*}
	\end{center}
	\vspace{-0.05in}
\end{table}

\setlength{\tabcolsep}{2pt}
\begin{table}[t]
	\scriptsize
	\begin{center}
		\caption{Comparison results of the DiMP pipeline between without and with our transformer on 7 tracking benchmarks. We compute the relative gain in the VOT2018, while in the rest datasets, we exhibit the absolute gain.} \label{table:dimp_ablation}	
		\vspace{+0.05in} 
		\begin{tabular*}{8.2 cm} {@{\extracolsep{\fill}}lccc}
			\hline
			~Dataset &DiMP Baseline &{\bf TrDiMP (Ours)} &$\Delta$\\
			\hline
			~Need for Speed \cite{NFSdataset} (AUC) &64.7 &\bf 66.5  &\bf 1.8\%$ \uparrow $\\
			~OTB-2015 \cite{OTB-2015} (AUC)  &70.1 &\bf 71.1 &\bf 1.0\%$ \uparrow $\\
			~UAV123 \cite{UAV123} (AUC)  &67.2 &\bf 67.5  &\bf 0.3\%$ \uparrow $\\
			~LaSOT \cite{LaSOT} (AUC)  &63.0 &\bf 63.9 &\bf 0.9\%$ \uparrow $\\
			~GOT-10k \cite{GOT10k} (AO) &66.7 &\bf 68.8  &\bf 2.1\%$ \uparrow $\\
			~TrackingNet \cite{2018trackingnet} (Success) &78.1 &\bf 78.4  &\bf 0.3\%$ \uparrow $\\
			~VOT2018 \cite{VOT2018} (EAO)  &0.446 &\bf 0.462 &\bf 3.6\%$ \uparrow $\\
			\hline
			~Tracking Speed (FPS) &\bf 30 &26  &4 FPS $ \downarrow $\\
			\hline
		\end{tabular*}
	\end{center}
	\vspace{-0.1in}
\end{table}

\section{Visualization}

\subsection{Attention Visualization}

\begin{figure}
	\centering
	\includegraphics[width=8.5cm]{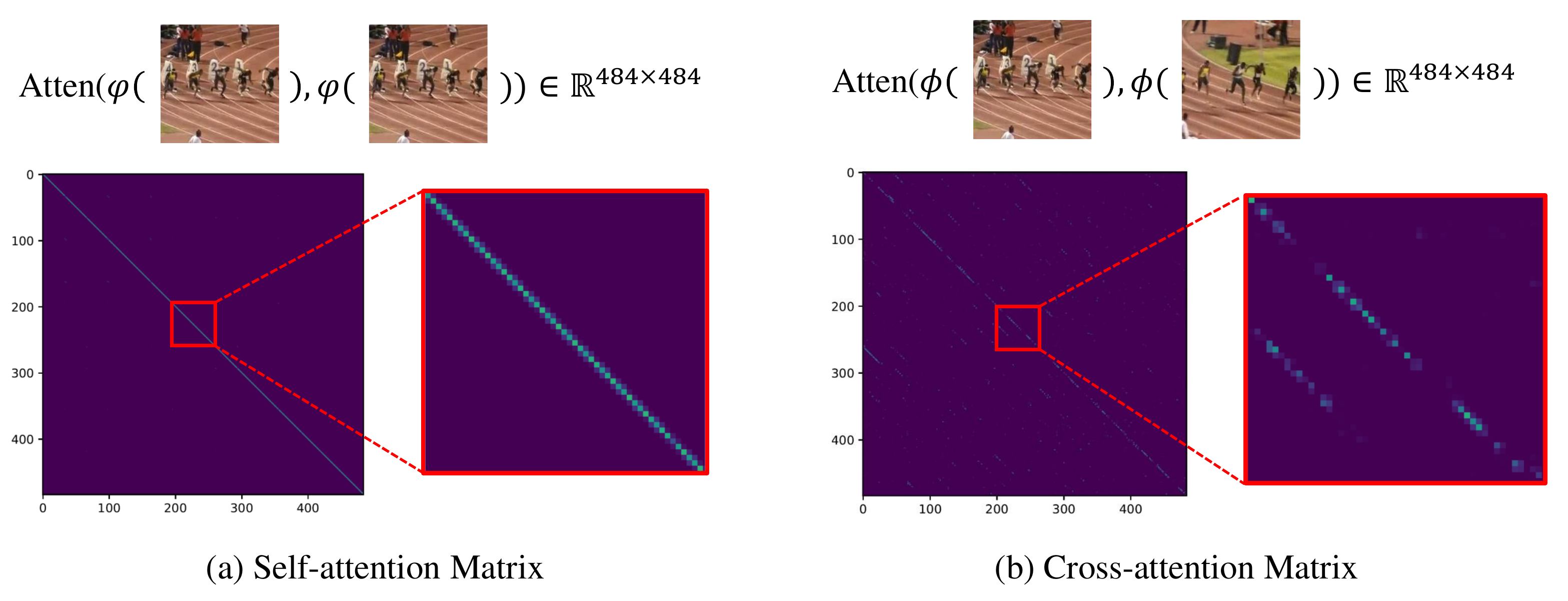}
	\caption{Visualization of the attention maps of the encoder (self-attention block) and decoder (cross-attention block).}
	\label{fig:atten_vis}
\vspace{-0.0in}
\end{figure}

As shown in Figure \ref{fig:atten_vis} (a), after self-attention, the pixels get some minor weights from their neighboring pixels to reinforce themselves. 
In the decoding process, as shown in Figure \ref{fig:atten_vis} (b), the cross-attention matrice between two different patches is sparse, which means the query seeks several most correlated keys to propagate the context.
After Softmax, the attention weights are not averaged by the similar athletes in {\tt Bolt2} sequence, which illustrates our attention block can discriminate the distractors to some extent. Benefiting such (feature/mask) propagations, the tracking responses are accurate, as shown in Figure~\ref{fig:vis}.

\subsection{Response Visualization}

In Figure~\ref{fig:vis}, we exhibit more detailed visualization results of our tracking framework.
From Figure~\ref{fig:vis} (second column), we can observe that our baseline (\emph{i.e.,} DiMP \cite{DiMP}) tends to be misled by distracting objects in the challenging scenarios.
By adopting the feature transformation mechanism (third column in Figure~\ref{fig:vis}), the target representations in the search region are effectively reinforced, which facilitates the object searching process.
Therefore, the response values of the background regions are largely restrained.
The mask transformation block propagates the spatial attentions from previous templates to the current search region, which also effectively suppresses the background objects (fourth column in Figure~\ref{fig:vis}).
Finally, our complete transformer architecture combines both feature and mask transformations, and the final response maps (last column in Figure~\ref{fig:vis}) are more robust for object tracking.

\begin{figure}
	\centering
	\includegraphics[width=8.5cm]{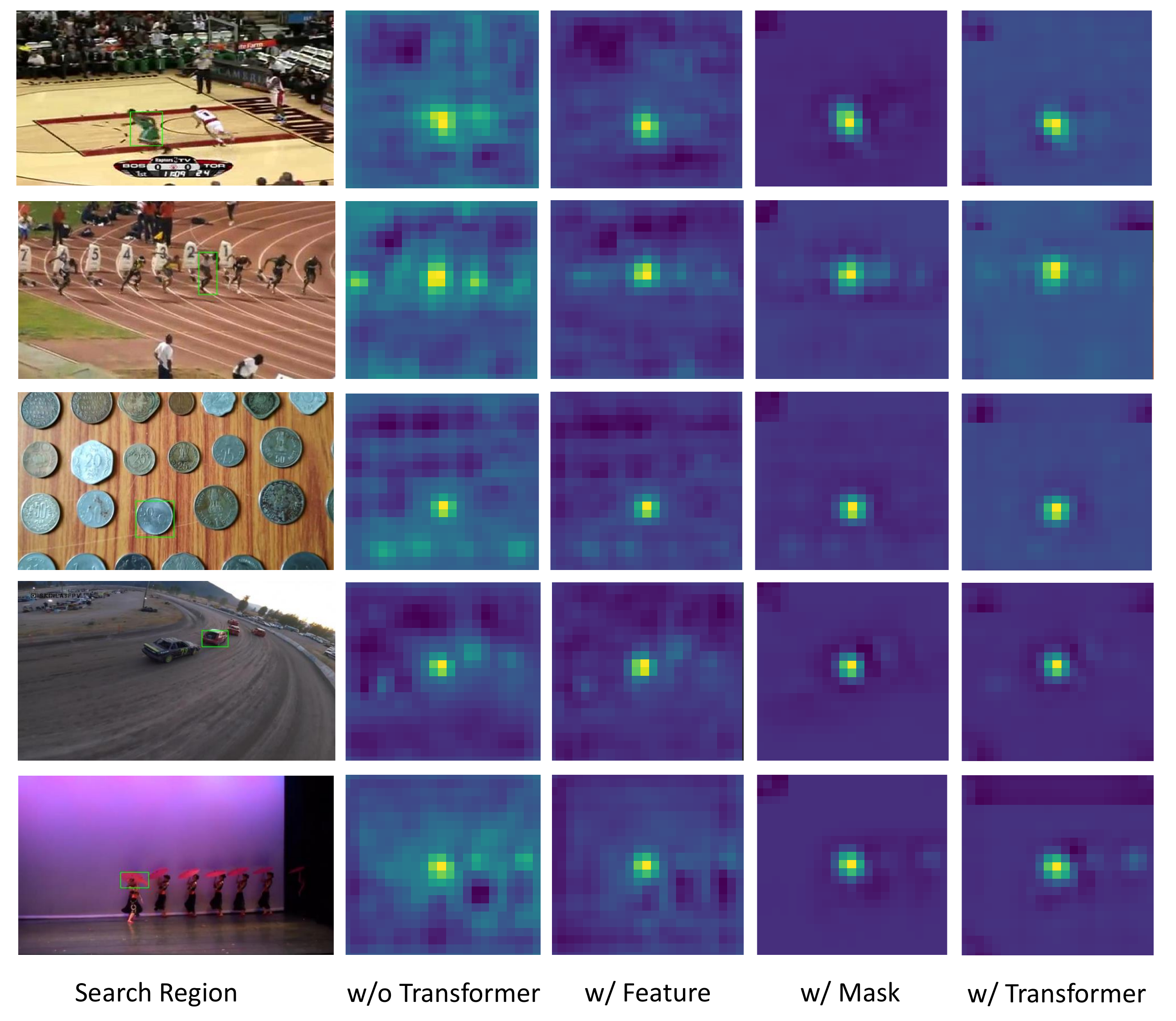}
	\caption{Visualization of the tracking response maps of DiMP baseline \cite{DiMP}. The ``w/o Transformer" denotes the baseline approach DiMP \cite{DiMP}. The ``w/ Feature" denotes the baseline with a feature propagation based transformer. The ``w/ Mask" represents the baseline with a mask propagation based transformer. Finally, the ``w/ Transformer" is our complete transformer-assisted tracker, \emph{i.e.,} TrDiMP. Our proposed components (feature and mask transformations) effectively suppress the background responses. }
	\label{fig:vis} 
\vspace{-0.05in}
\end{figure}

\section{Results on VOT2019}
VOT2019 \cite{VOT2019} is a recently released challenging benchmark, which replaces 12 easy videos in VOT2018 \cite{VOT2018} by 12 more difficult videos. 
%
%
We compare our approach with some top-performing approaches on VOT2019.
Table \ref{table:vot} shows the accuracy, robustness, and EAO scores of different trackers.
Compared with DiMP-50, our TrDiMP shows similar tracking accuracy but exhibits a much lower failure rate (\emph{i.e.,} robustness score).
Compared with other recent deep trackers with the ResNet-50 backbone, our TrDiMP significantly surpasses them such as SiamRPN++, SiamDW \cite{deeperwiderSiamFC} and SiamMask \cite{SiamMask} by a considerable margin.
The VOT2019 challenge winner (\emph{i.e.,} DRNet) shows an EAO score of 0.395 \cite{VOT2019}.
Overall, the proposed TrDiMP outperforms the current top-performing trackers with a promising EAO score of 0.397.

\setlength{\tabcolsep}{2pt}
\begin{table}[t]
	\scriptsize
	\begin{center}
		\caption{The accuracy (A), robustness (R), and expected average overlap (EAO) of state-of-the-art methods on the VOT-2019 \cite{VOT2019}.} \label{table:vot}	
		\vspace{+0.05in} 
		\begin{tabular*}{8.2 cm} {@{\extracolsep{\fill}}lccccccc}
			\hline
			&SPM &SiamRPN++ &SiamMask &ATOM  &SiamDW &DiMP-50 &{\bf TrDiMP} \\
			&\cite{SPM} &\cite{siamrpn++} &\cite{SiamMask} &\cite{ATOM}  &\cite{deeperwiderSiamFC} &\cite{DiMP} &Ours \\
			\hline  
			~A ($ \uparrow $)  &0.577 &0.599 &0.594  &{\bf \color{red} 0.603} &{\bf \color{blue} 0.600} &0.594 & 0.598  \\
			~R ($ \downarrow $) &0.507 &0.482 &0.461  &0.411 &0.467 &{\bf \color{blue} 0.278} & {\bf \color{red} 0.231}  \\
			~EAO ($ \uparrow $)  &0.275 &0.285 &0.287 &0.292   &0.299 &{\bf \color{blue} 0.379} & {\bf \color{red} 0.397}  \\
			\hline
		\end{tabular*}
	\end{center}
\vspace{-0.2in}
\end{table}

\section{Failure Case}

When the target object is occluded or invisible, the cross attention maps between the current frame and historic templates are inaccurate. Therefore, our framework struggles to handle the heavy occlusion (\emph{e.g.,} Figure \ref{fig:failure}) or out-of-view. 
Another potential limitation of our work is the high computational memory of the attention matrix, which is also a common issue in the transformer.
%

{\makeatletter\def\@captype{table}\makeatother
	\begin{figure}[h]
		\centering
		\includegraphics[width=8.5cm]{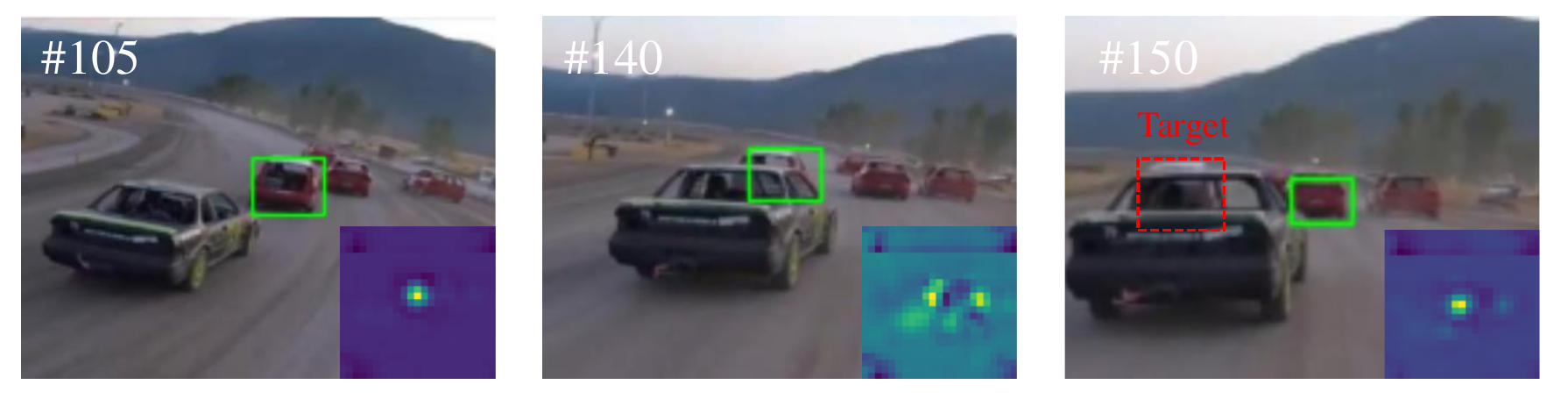}
		\caption{Failure case. TrDiMP fails to track the occluded {target}.}
		\label{fig:failure} 
	\end{figure}
\vspace{-0.1in}
}

\section{Attribute Analysis}

\begin{figure*}
	\centering
	\includegraphics[width=4.9cm]{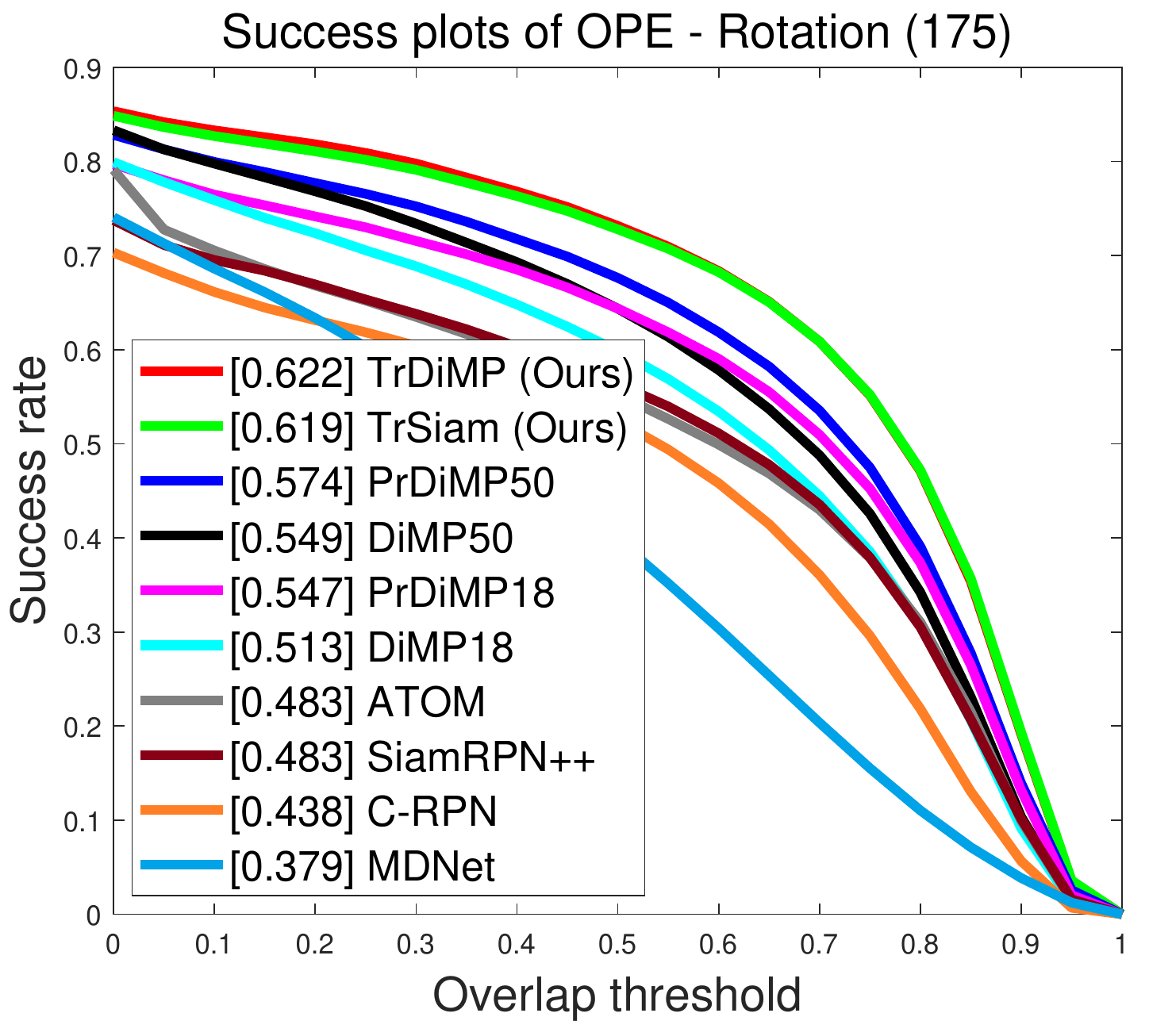}
	\includegraphics[width=4.9cm]{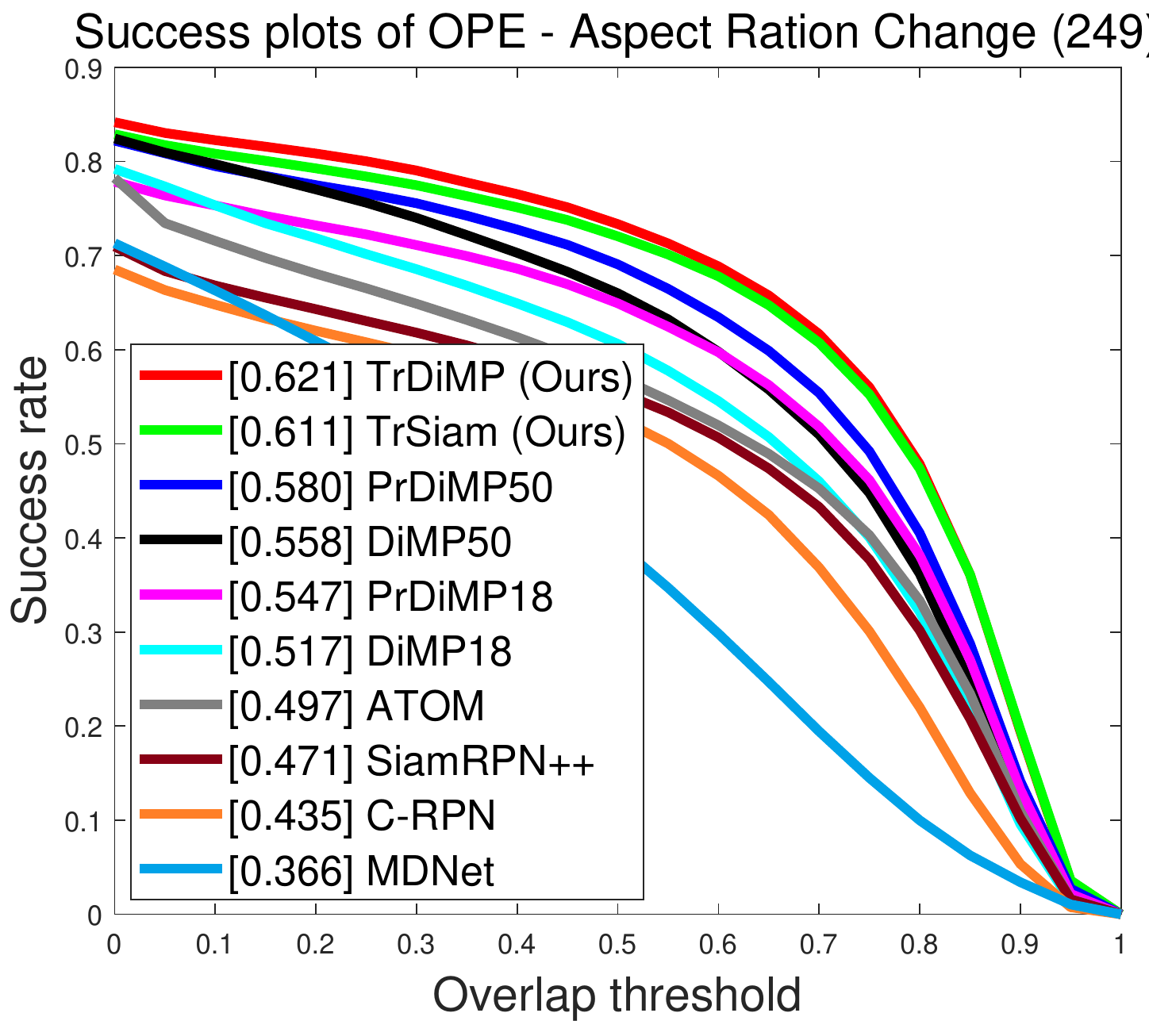}
	\includegraphics[width=4.9cm]{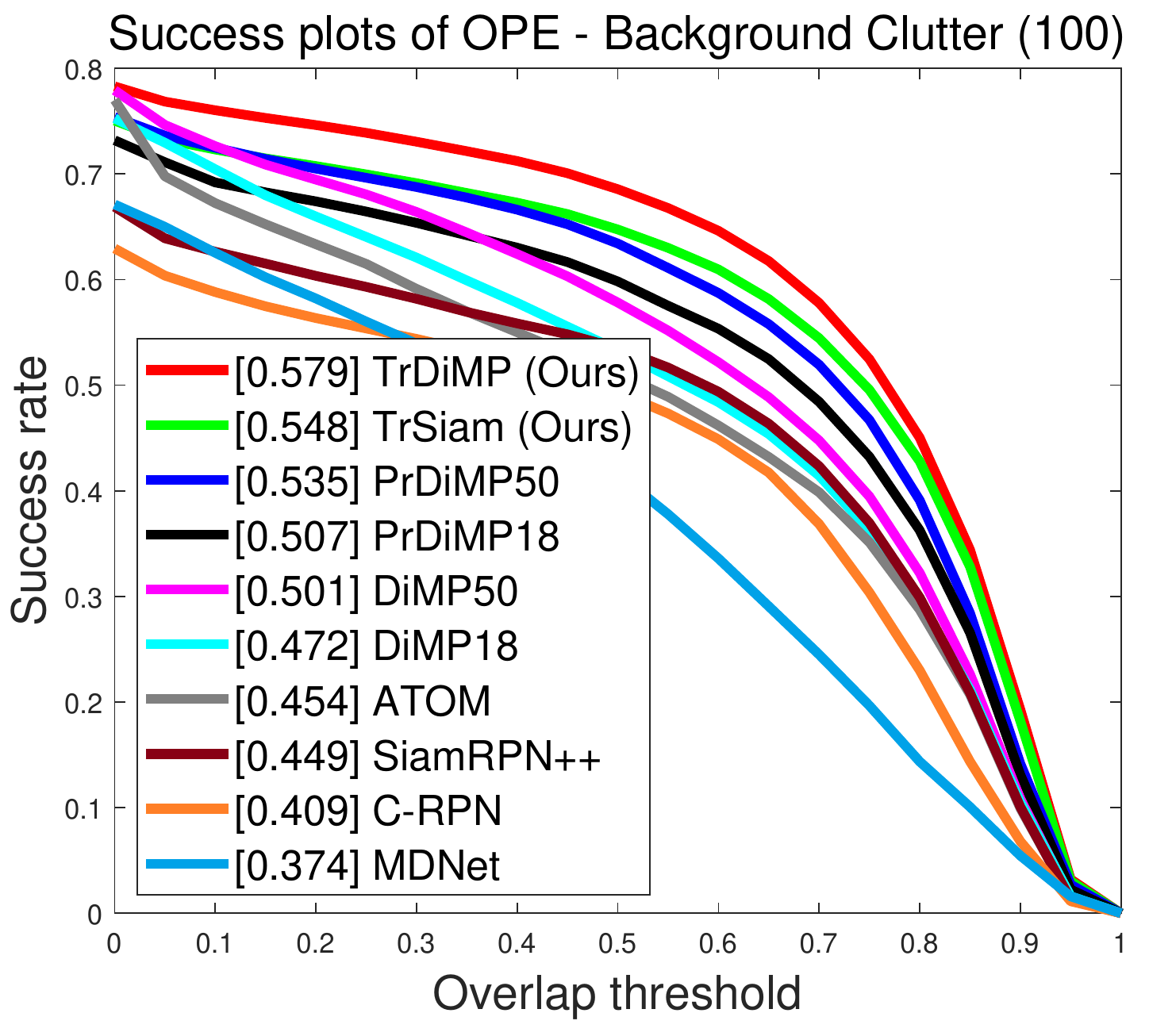}
	\includegraphics[width=4.9cm]{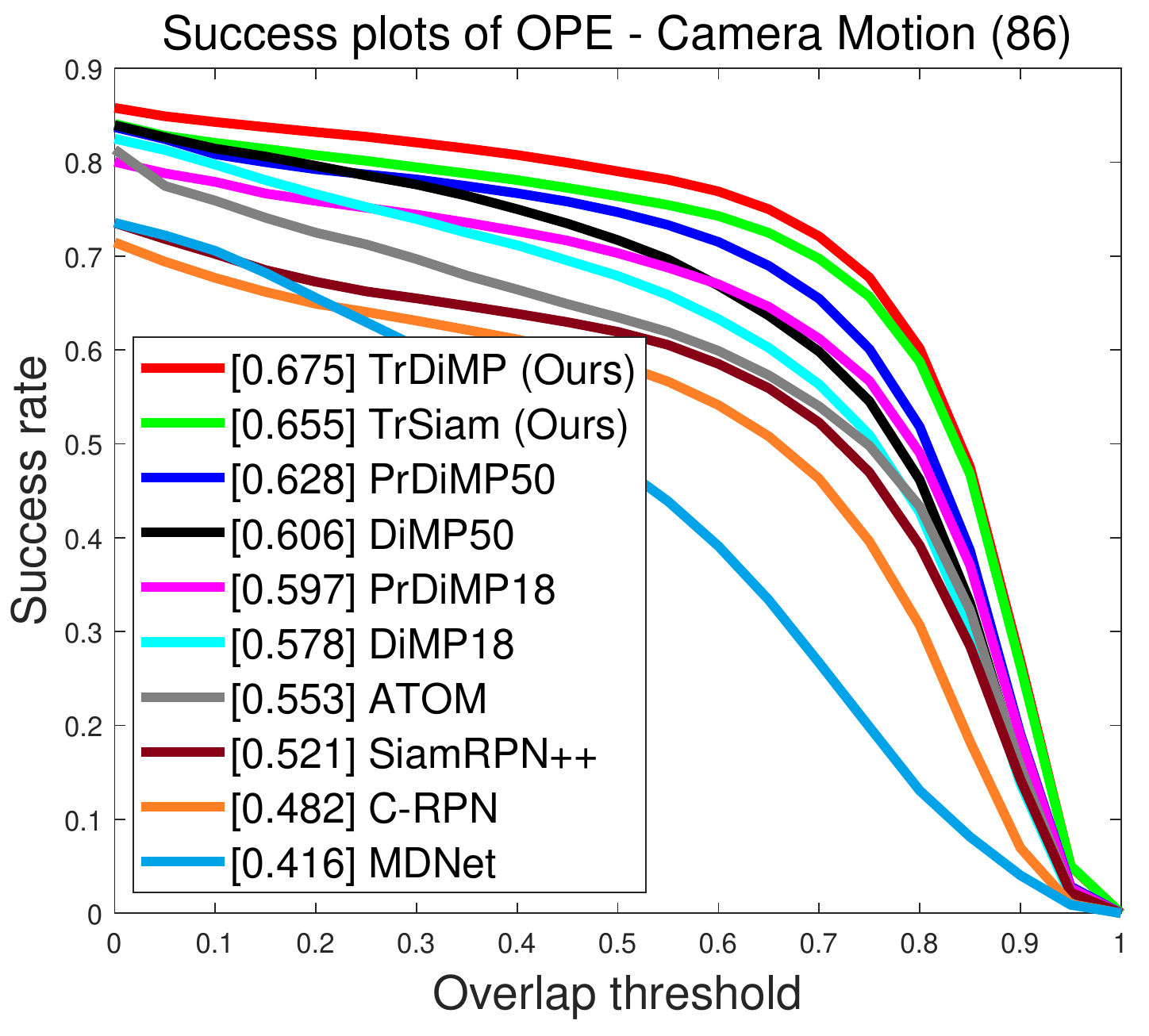}
	\includegraphics[width=4.9cm]{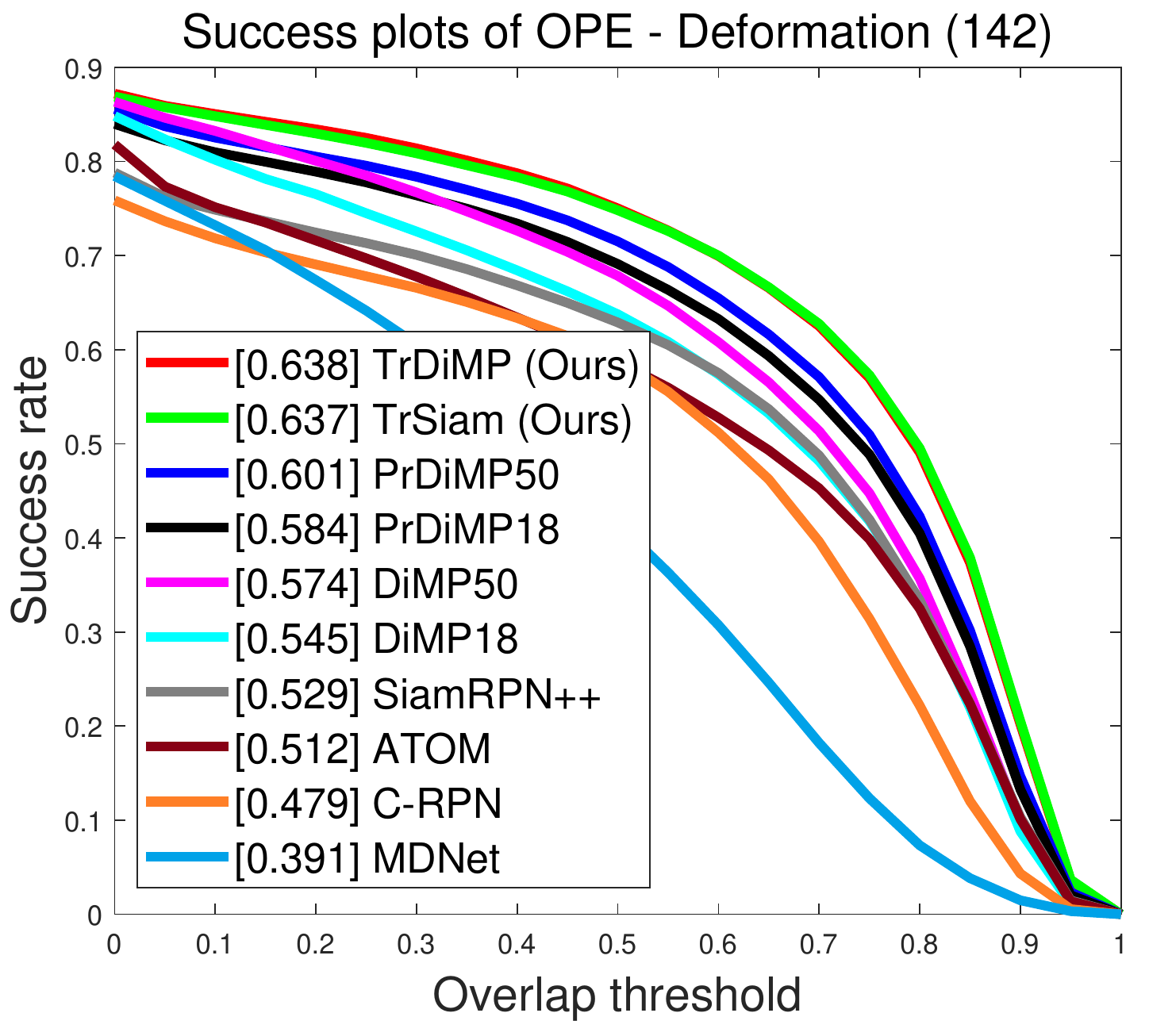}
	\includegraphics[width=4.9cm]{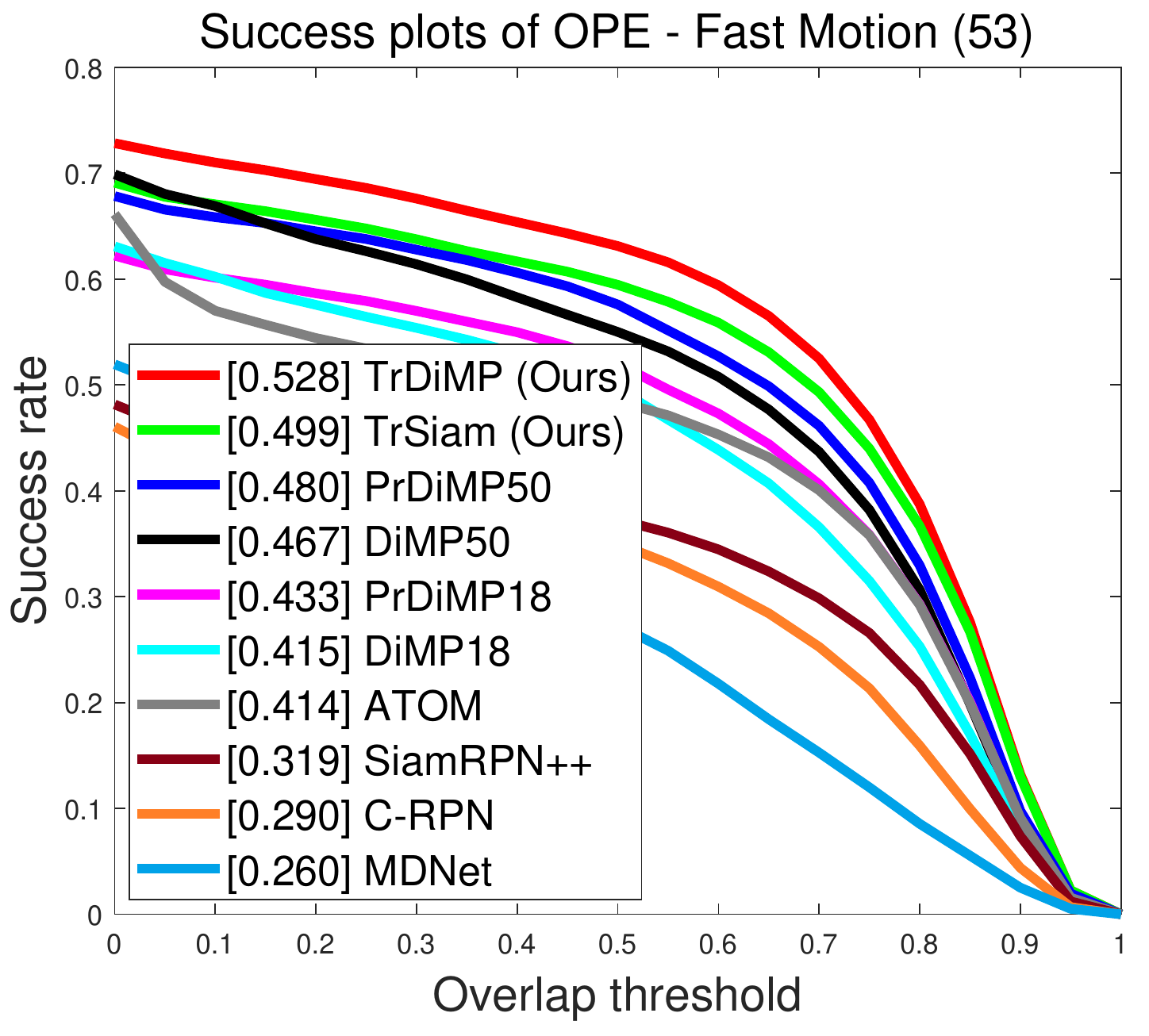}
	\includegraphics[width=4.9cm]{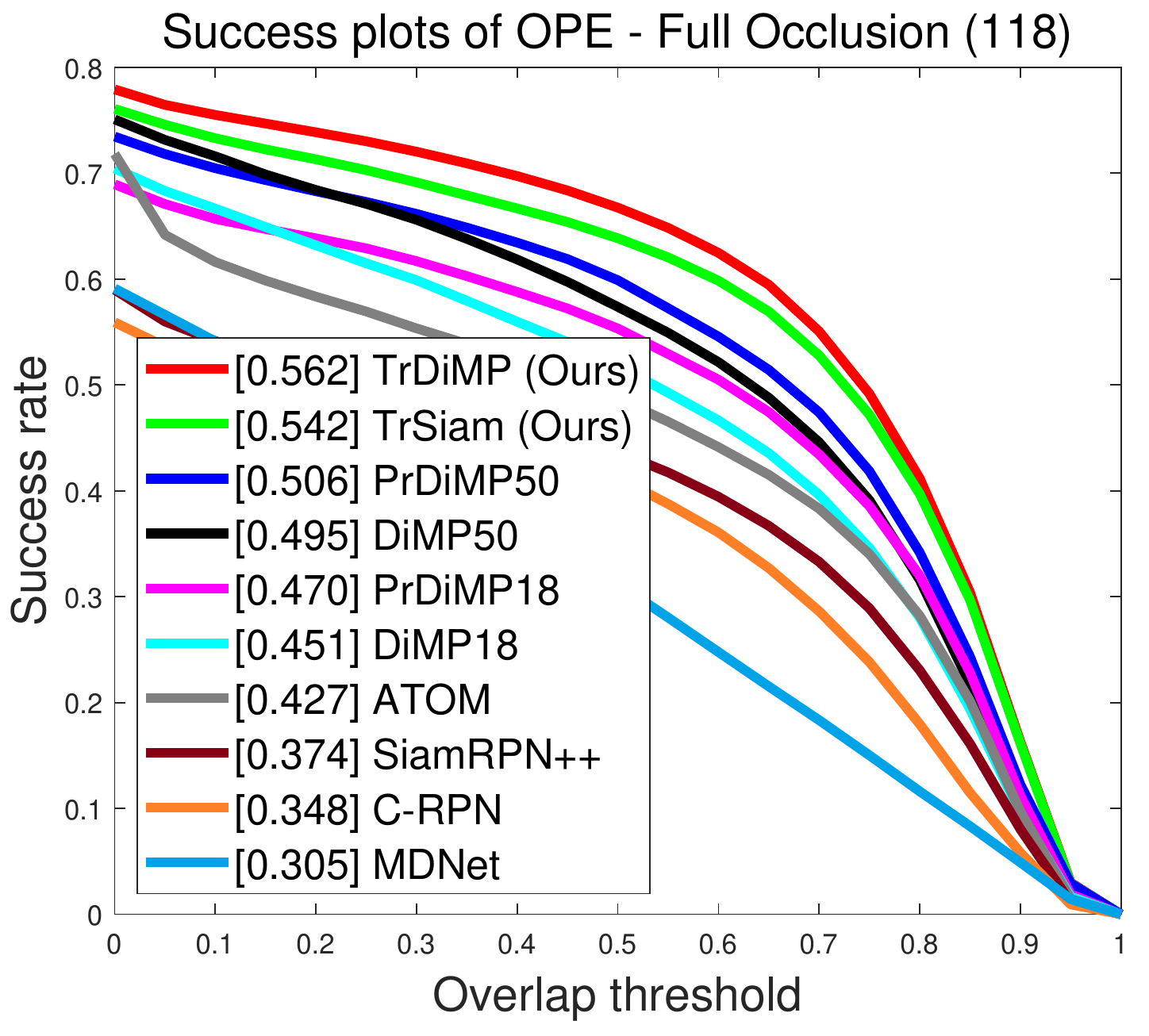}
	\includegraphics[width=4.9cm]{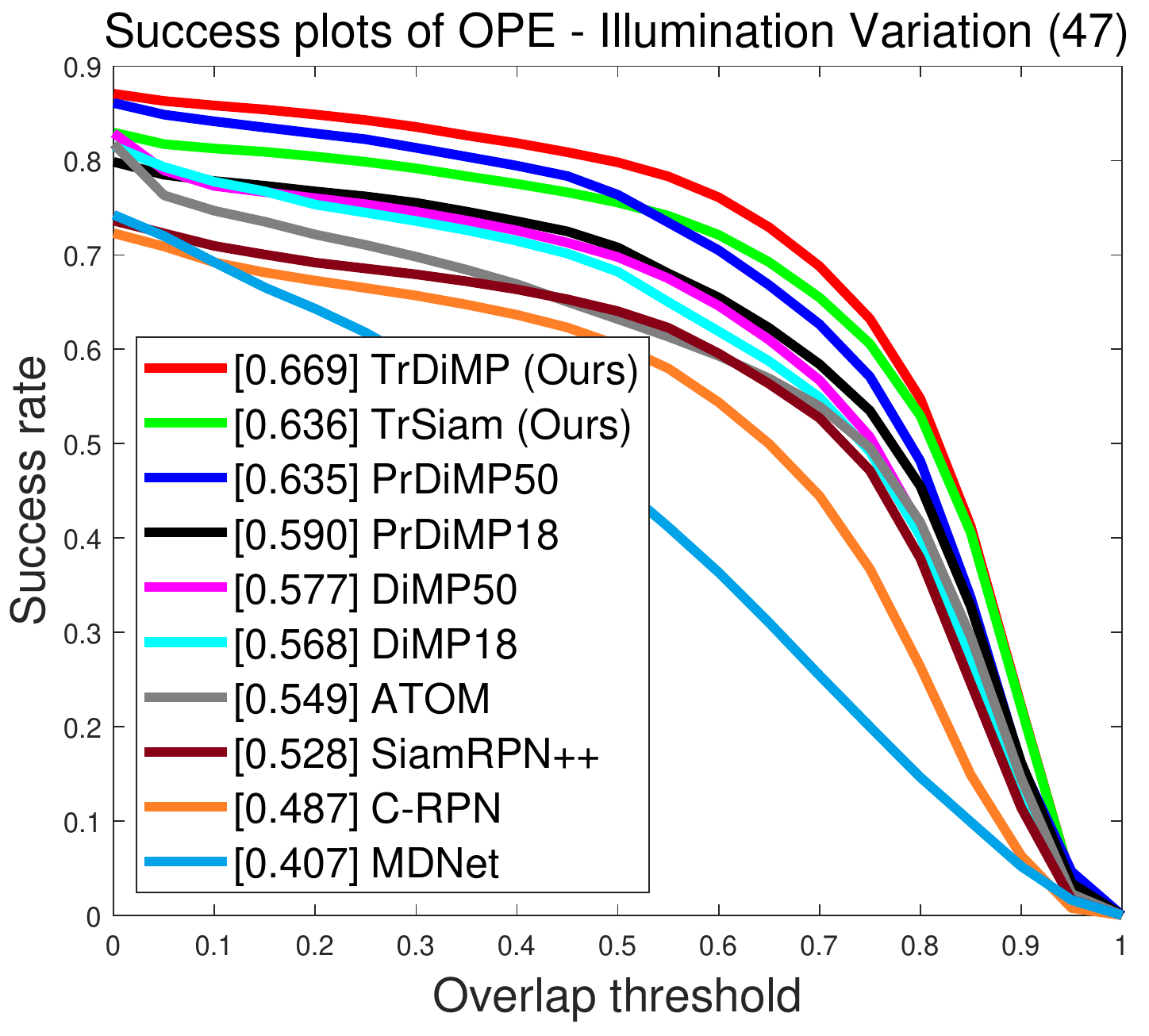}
	\includegraphics[width=4.9cm]{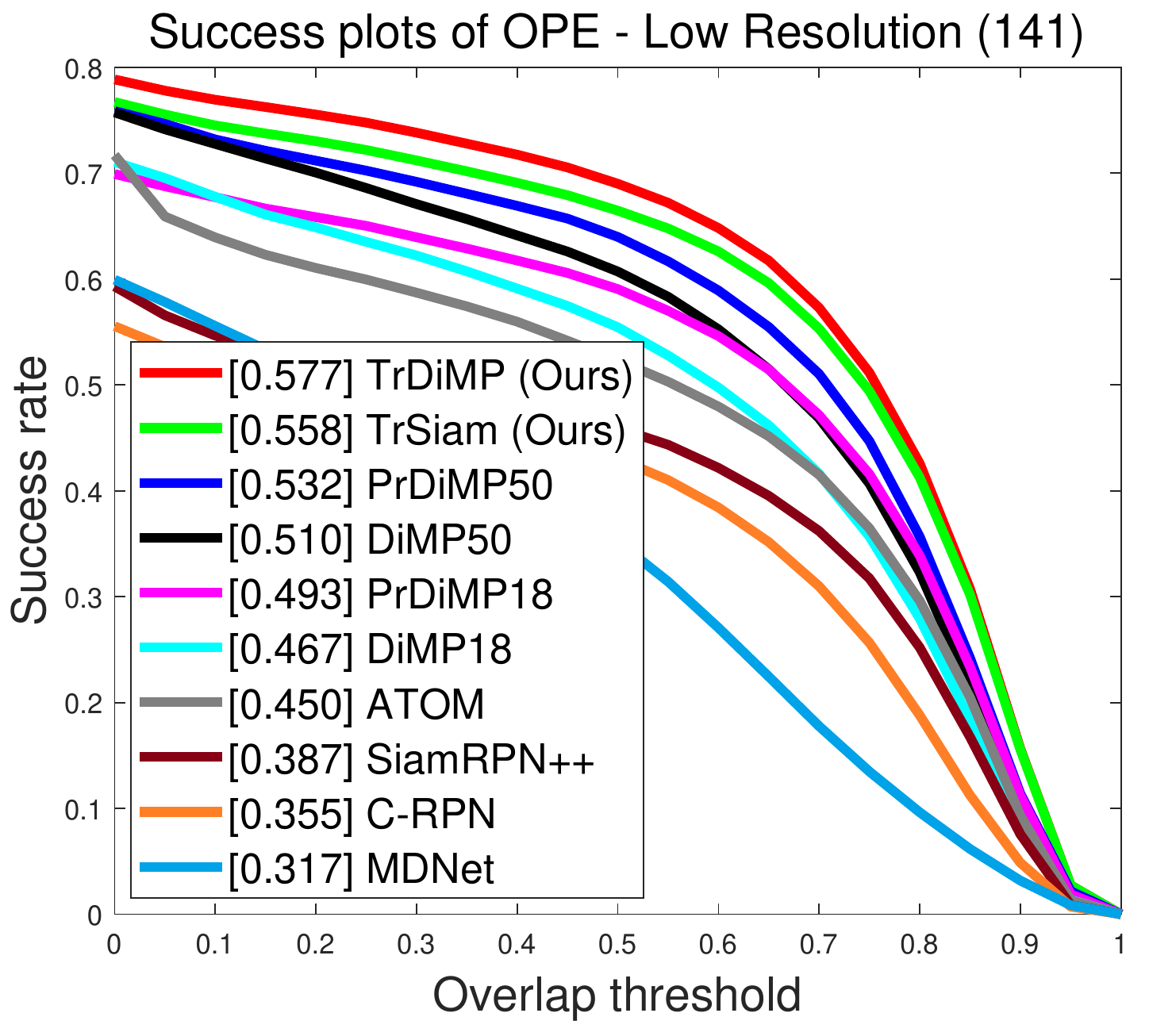}
	\includegraphics[width=4.9cm]{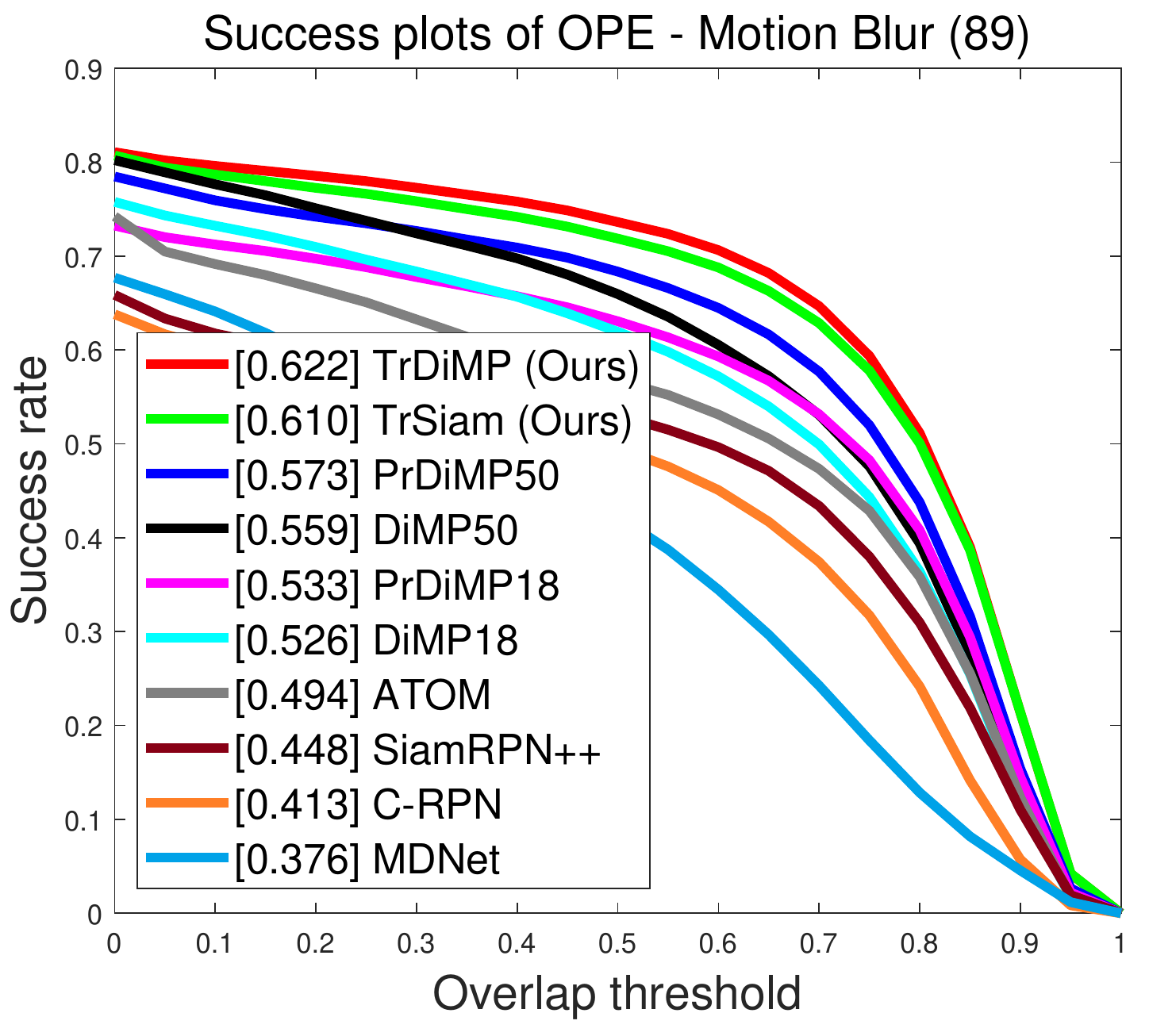}
	\includegraphics[width=4.9cm]{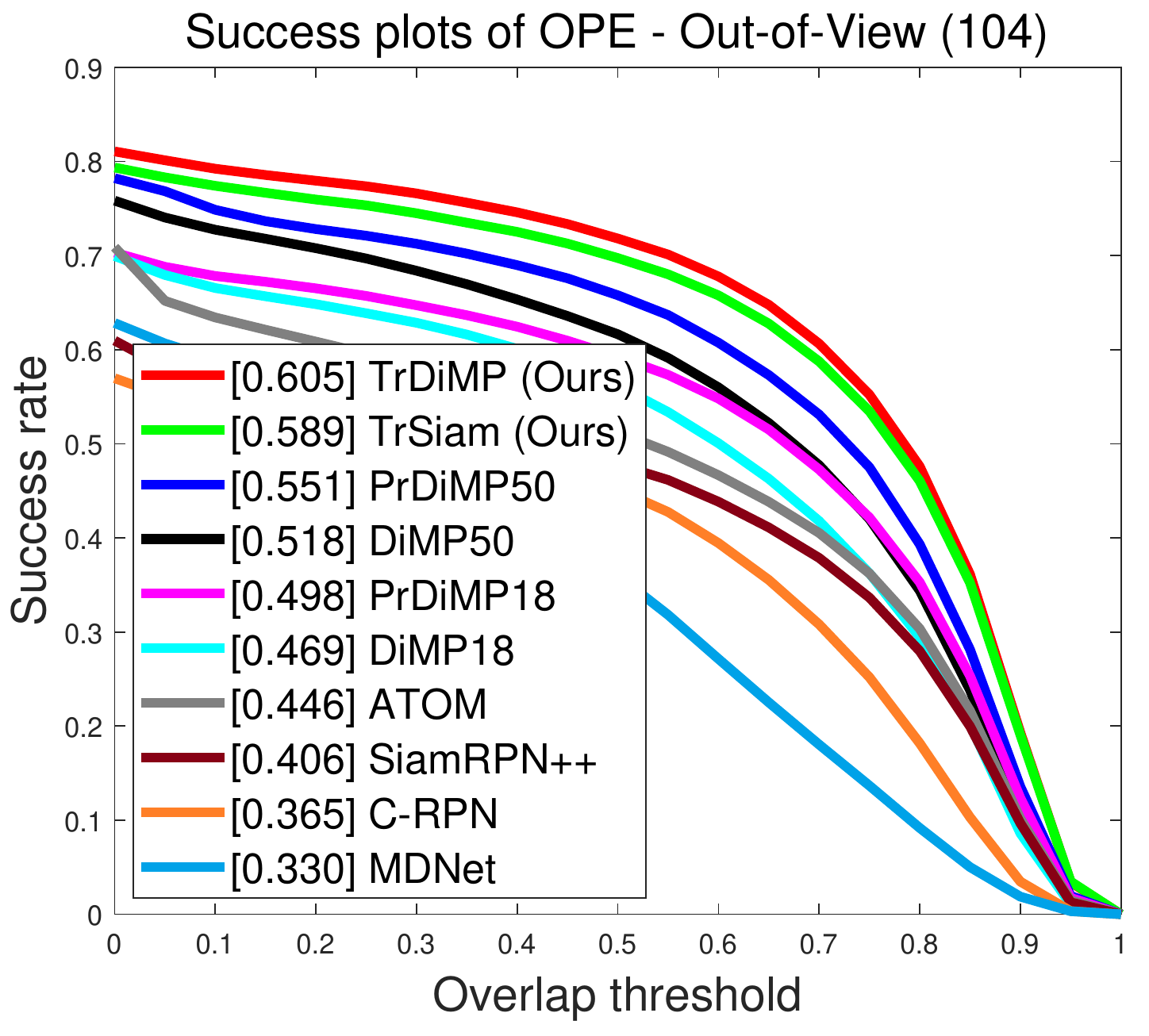}
	\includegraphics[width=4.9cm]{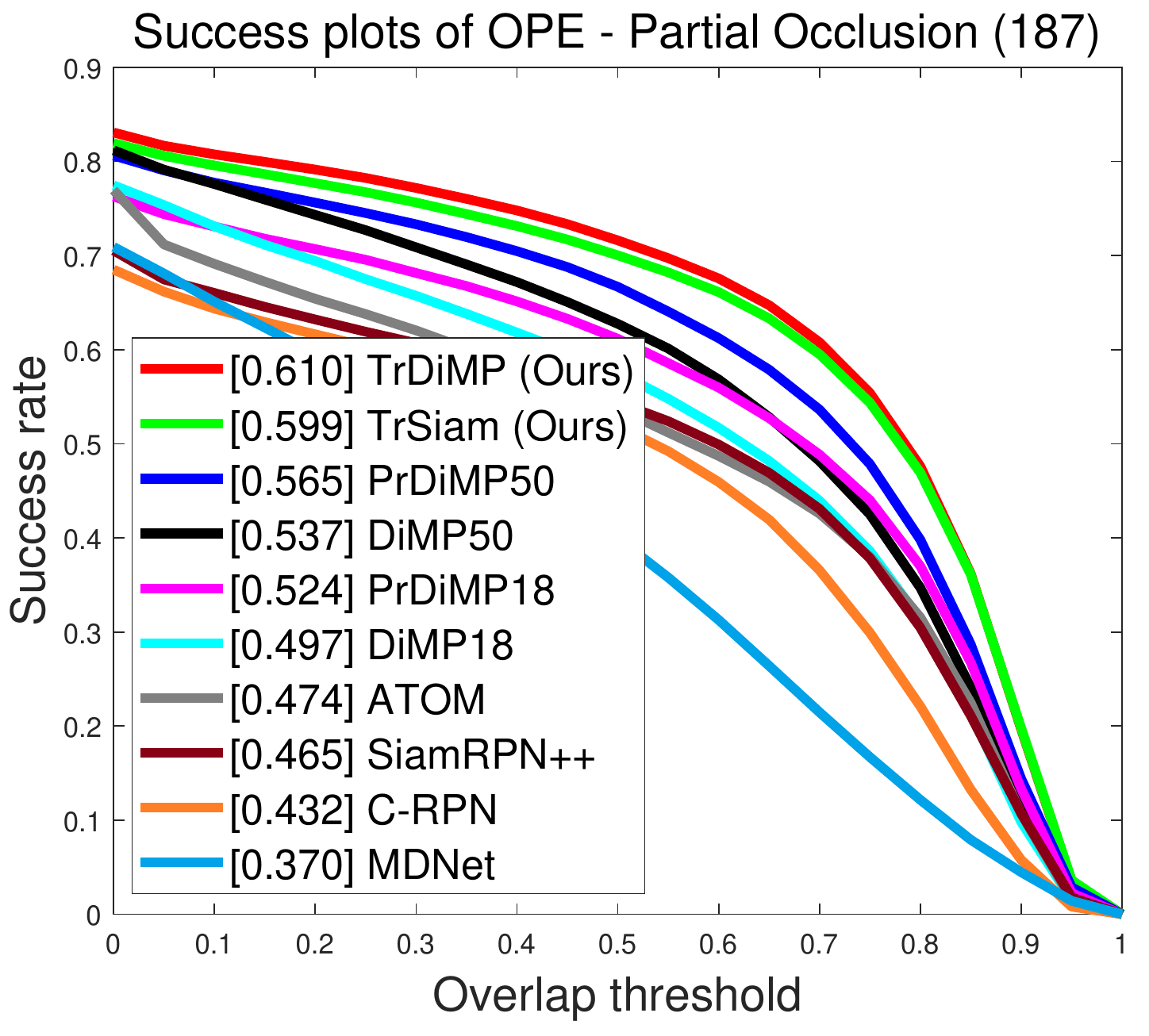}
	\includegraphics[width=4.9cm]{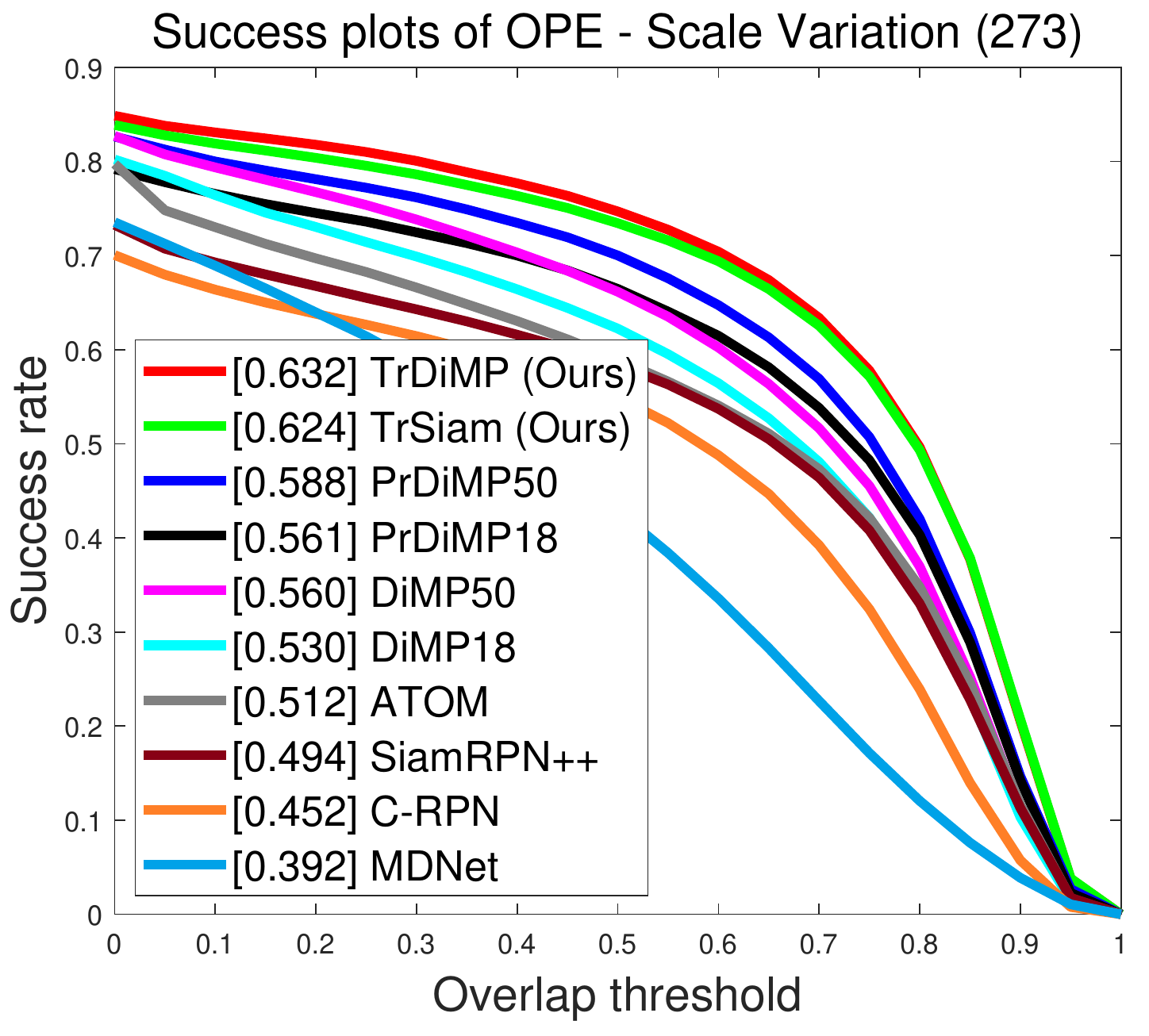}
	\includegraphics[width=4.9cm]{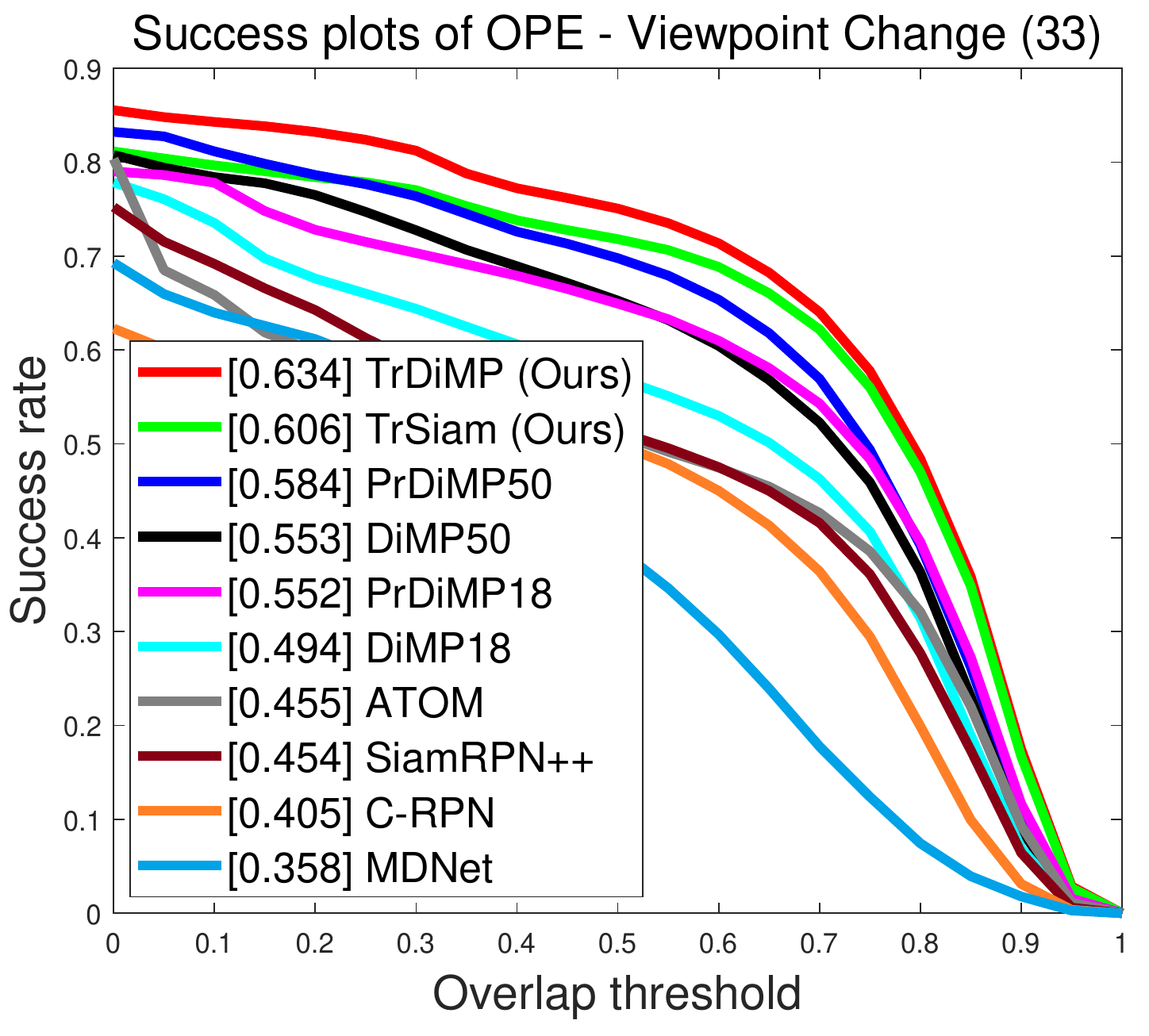}
	\caption{Attribute-based evaluation on the LaSOT benchmark \cite{LaSOT}. The legend shows the AUC scores of the success plots.}
	\label{fig:lasot-attribute} 
\end{figure*}

Finally, in Figure~\ref{fig:lasot-attribute}, we provide the attribute evaluation on the LaSOT \cite{LaSOT} benchmark.
On the LaSOT, our approaches show good results in various scenarios such as motion blur, background clutter, low resolution, and viewpoint change.
%
%
As shown in Table~\ref{table:siam_ablation}, with the proposed transformer, our TrSiam outperforms its baseline by  3.3\% AUC.
It should be noted that our simple TrSiam does not adopt complex online model optimization techniques, which is more efficient than the recent approaches such as DiMP \cite{DiMP} and PrDiMP \cite{PrDiMP}.

\end{document}